\documentclass[sigconf]{acmart}

\usepackage[T1]{fontenc}

\usepackage{booktabs} \usepackage{microtype}
\usepackage{graphicx}
\usepackage{amsmath,amsthm}
\usepackage{hyperref}
\usepackage{float}

\usepackage{url}
\usepackage{breakcites}
\usepackage{subcaption}
\usepackage{multirow}
\usepackage{makecell}
\usepackage{pifont}
\usepackage{breakcites}
\usepackage{balance}

\usepackage{Definitions}

\copyrightyear{2018} 
\acmYear{2018} 
\setcopyright{rightsretained} 
\acmConference[KDD '18]{The 24th ACM SIGKDD International Conference on Knowledge Discovery \& Data Mining}{August 19--23, 2018}{London, United Kingdom}
\acmBooktitle{KDD '18: The 24th ACM SIGKDD International Conference on Knowledge Discovery \& Data Mining, August 19--23, 2018, London, United Kingdom}\acmDOI{10.1145/3219819.3220046}
\acmISBN{978-1-4503-5552-0/18/08}

\begin{document}
\title{A Unified Approach to Quantifying Algorithmic Unfairness: Measuring Individual \& Group Unfairness via Inequality Indices}
\renewcommand{\shorttitle}{A Unified Approach to Quantifying Algorithmic Unfairness}

\author{Till Speicher$^{1,*}$, Hoda Heidari$^{2,*}$, Nina Grgic-Hlaca$^1$, \\Krishna P. Gummadi$^1$, Adish Singla$^1$, Adrian Weller$^{3,4}$, Muhammad Bilal Zafar$^1$}
\affiliation{  \institution{$^1$MPI-SWS, $^2$ETH Zurich, $^3$University of Cambridge, $^4$The Alan Turing Institute}
}
\renewcommand{\shortauthors}{Speicher et al.}

\fancyhead{}

\begin{abstract}
Discrimination via algorithmic decision making has received considerable attention. 
Prior work largely focuses on defining \emph{conditions} for fairness, but does not define satisfactory \emph{measures} of algorithmic unfairness. In this paper, we focus on the following question: Given two unfair algorithms, how should we determine which of the two is more unfair?
Our core idea is to use existing inequality indices from economics to measure how unequally the outcomes of an algorithm benefit different individuals or groups in a population.
Our work offers a justified and general framework to compare and contrast the (un)fairness of algorithmic predictors. This unifying approach enables us to quantify unfairness both at the individual and the group level. Further, our work reveals overlooked tradeoffs between different fairness notions: using our proposed measures, the \emph{overall individual-level} unfairness of an algorithm can be decomposed into a \emph{between-group} and a \emph{within-group} component. Earlier methods are typically designed to tackle only between-group unfairness, which may be justified for legal or other reasons. However, we demonstrate that minimizing exclusively the between-group component may, in fact, increase the within-group, and hence the overall unfairness. We characterize and illustrate the tradeoffs between our measures of (un)fairness and the prediction accuracy.
\end{abstract} 

\maketitle
{\let\thefootnote\relax\footnote{{${}^*$ Authors contributed equally to this work.}}}
\setcounter{footnote}{0}

\section {Introduction} \label{sec:intro}

As algorithmic decision making systems are increasingly used in life-affecting scenarios such as criminal risk
prediction~\cite{propublica_story,berk2017fairness} and credit risk
assessments~\cite{history-of-credit}, concerns have risen about the 
potential unfairness of these decisions to certain social groups or
individuals~\cite{barocas2016big,bigdatawhitehouse,salvatore_survey}.
In response, a number of recent works have proposed learning
mechanisms for fair decision making by imposing additional constraints
or {\it
  conditions}~\cite{feldman_kdd15,zafar_fairness,zafar_dmt,hardt_nips16,dwork2012fairness,joseph2016fairness}.

In this paper, we focus on a simple yet foundational question about
unfairness of algorithms: {\it Given two unfair algorithms, how should
  we determine which of the two is more unfair?}  Prior works on
algorithmic fairness largely focus on formally defining {\it
  conditions} for fairness, but do not precisely define suitable {\it measures}
for unfairness. That is, they can answer the binary question: {\it is
  an algorithm fair or unfair?}, but do not have a principled way to answer the nuanced question: {\it if
  an algorithm is unfair, how unfair is it?}

Figure~\ref{fig:comparing_clfs} illustrates
the questions we seek to answer through an example of two binary
classifiers $\mathcal{C}_1$ and $\mathcal{C}_2$, whose decisions
affect 10 individuals belonging to 3 different groups.
The figure shows that both $\mathcal{C}_1$ and $\mathcal{C}_2$ yield
unequal false positive/negative rates across the 3 groups and are thus
unfair at the level of groups---which set of unequal false
positive/negative rates ($\mathcal{C}_1$'s or $\mathcal{C}_2$'s) are
more unfair?  Similarly, both $\mathcal{C}_1$ and $\mathcal{C}_2$
violate our individual-level fairness condition for ``treating
individuals deserving similar outcomes similarly'', but do so in
different ways---whose violation of our individual fairness condition
is more unfair?

We argue that how we address the unfairness measurement question has
significant practical consequences. First, several studies have
observed that satisfying multiple fairness conditions at the same time
is infeasible~\cite{klein16,dimpact_fpr,Corbett-DaviesP17,kearns2017preventing}. Hence in practice, designers often need to
select the {\it least unfair} algorithm from a feasible set of unfair
algorithms. 
Second, when training fair learning models, practitioners face a
tradeoff between accuracy and
fairness~\cite{flores_propublica_reply,klein16}. These tradeoffs rely
on model-specific fairness measures (i.e., proxies chosen for
computational tractability) that do not {\it generalize} across
different models. Consequently, they cannot be used to compare
accuracy-unfairness tradeoffs of models trained using different fair
learning algorithms.
Finally, designers of fair
learning models make a number of {\it ad hoc or implicit} choices about
fairness measures without explicit justification; for instance,
it is unclear why in many previous works~\cite{zafar_fairness,zafar_dmt,hardt_nips16,dwork2012fairness}, the relative
sizes of the groups in the population are not considered in estimating unfairness---even though these quantities matter when estimating accuracy. 

In this paper, we propose to quantify unfairness using {\it inequality
  indices} that have been extensively studied in economics and social
welfare~\cite{atkinson1970measurement,cowell1981additivity,kakwani1980class}. Traditionally, inequality indices such as
Coefficient of Variation~\cite{abdi2010coefficient}, Gini~\cite{gini1912variabilita,bellu2006inequality}, Atkinson~\cite{atkinson1970measurement},
Hoover~\cite{long1997hoover}, and Theil~\cite{theil1967}, have been proposed to quantify how
unequally incomes are distributed across individuals and groups in a
population. Our interest in using these indices is rooted in the {\it
  well-justified} axiomatic basis for their designs. Specifically, we
argue that many axioms satisfied by inequality indices such as {\it
  anonymity}, {\it population invariance}, {\it progressive transfer
  preference}, and {\it subgroup decomposability} are appealing properties
for unfairness measures to satisfy. Thus, inequality indices are
naturally well-suited as measures for algorithmic unfairness.

Our core idea is to use existing inequality indices in order to
measure {\it how {\bf unequally} the outcomes of an algorithm {\bf benefit}
  different individuals or groups in a population}. This requires
us to define a benefit function  that maps the
algorithmic output for each individual to a non-negative real
number.   By adapting the benefit function according to the
desired fairness condition,
we show that inequality indices can be applied generally to
quantify unfairness across all the proposed fairness conditions shown in
Figure~\ref{fig:clf_fairness}. Since we quantify inequality of
algorithmic outcomes, our measure is independent of the specifics of
any learning model and can be used to compare unfairness of different
algorithms.

We consider a family of inequality indices called {\it
  generalized entropy indices}, which includes Coefficient of
Variation and Theil index as special cases. Generalized entropy indices have a useful
property called {\it subgroup decomposability}.
For any division of the population into a set of non-overlapping
groups, the property guarantees that our unfairness measure over the
entire population can be decomposed as the sum of a {\it between-group}
unfairness component (computed imagining that all individuals in a
group receive the group's mean benefit) and a {\it within-group}
unfairness component (computed as a weighted sum of inequality in
benefits received by individuals within each group). Thus, inequality
indices not only offer a {\it unifying} approach to quantifying
unfairness at the levels of both individuals and groups, but they also
reveal previously overlooked {\it tradeoffs} between individual-level and
group-level fairness.

Further, the decomposition enables us to:  (i) quantify how unfair an algorithm is along various sensitive
attribute-based groups within a population (e.g., groups based on
race, gender or age) and (ii) account for the ``gerrymandered''
unfairness affecting structured subgroups constructed from
``intersecting'' the sensitive attribute-groups (e.g., groups like
young white women or old black men) \cite{kearns2017preventing}. 
Our empirical evaluations show
that existing fair learning methods~\cite{zafar_dmt,hardt_nips16}, while successful in
eliminating between-group unfairness, (a) may be targeting only a small
fraction of the overall unfairness in the decision making algorithms
and (b) can result in an increase in within-group unfairness, which
paradoxically can lead to training algorithms whose overall unfairness
is worse than those trained using traditional learning methods.

To summarize the contributions of this paper: (i) we propose
inequality-indices based unfairness measures that offer a
justified and generalizable framework to compare the fairness of
a variety of algorithmic predictors against one another, (ii) we
theoretically characterize and empirically illustrate the
tradeoffs between individual fairness when measured using inequality
indices and the prediction accuracy, and (iii) we study the relationship between
individual- and group-level unfairness, showing that recently proposed learning
mechanisms for mitigating (between-)group unfairness can lead
to high within-group unfairness and consequently, high individual
unfairness.

\begin{figure}[ht]
\centering
\resizebox{\columnwidth}{!}{
\begin{tabular}{|c|c||c|c|c|c|c|c|c|c|c|c|}
\hline
\multicolumn{2}{ | c ||}{{\bf Individuals}} &
{\color{Red} ${i_1}$} &
{\color{Red} ${i_2}$} &
{\color{Green} ${i_3}$} &
{\color{Green} ${i_4}$} &
{\color{Green} ${i_5}$} &
{\color{Green} ${i_6}$} &
{\color{blue} ${i_7}$} &
{\color{blue} ${i_8}$} &
{\color{blue} ${i_9}$} &
{\color{blue} ${i_{10}}$} 
\\ \hline
\multicolumn{2}{ | c ||}{{\bf Groups}} &
{\color{Red} ${g_1}$} &
{\color{Red} ${g_1}$} &
{\color{Green} ${g_2}$} &
{\color{Green} ${g_2}$} &
{\color{Green} ${g_2}$} &
{\color{Green} ${g_2}$} &
{\color{blue} ${g_3}$} &
{\color{blue} ${g_3}$} &
{\color{blue} ${g_3}$} &
{\color{blue} ${g_3}$} 
\\ \hline
\multicolumn{2}{ | c ||}{\makecell{\bf True  Labels}} & 
1 & 0 & 0 & 1 & 0 & 0 & 1 & 0 & 1 & 1
\\ \hline
\multirow{2}{*}{ \makecell{\bf Predicted\\ \bf Labels} } & $\mathcal{C}_1$ &
1 & 0 & 0 & 0 & 1 & 1 & 1 & 0 & 1 & 0
\\ \cline{2-12}
& $\mathcal{C}_2$ &
0 & 1 & 1 & 0 & 0 & 0 & 0 & 1 & 1 & 1
\\ \hline
\end{tabular}
}
\\
\vspace{2mm}
\resizebox{\columnwidth}{!}{
\begin{tabular}{c| >{\color{Red}}c | >{\color{Green}} c | >{\color{blue}} c | c | >{\color{Red}} c | >{\color{Green}} c | >{\color{blue}} c | c |}
\cline{2-9}
& \multicolumn{4}{ c |}{$\mathcal{C}_1$} & \multicolumn{4}{ c |}{$\mathcal{C}_2$} \\
\cline{2-9}
& $\mathbf{g_1}$ & $\mathbf{g_2}$ & $\mathbf{g_3}$& {\bf Fair?} &$\mathbf{g_1}$ & $\mathbf{g_2}$ & $\mathbf{g_3}$ & {\bf Fair?}\\
\hline
\multicolumn{1}{ |c ||}{\bf FPR} & 
0.00 & 0.67 & 0.00 &
\xmark &
1.00 & 0.33 & 1.00 
& \xmark
\\ \hline
\multicolumn{1}{ |c ||}{\bf FNR} & 
0.00 & 1.00 & 0.33 &
\xmark &
1.00 & 1.00 & 0.33 
& \xmark
\\ \hline
\multicolumn{1}{ |c ||}{\bf FDR} & 
0.00 & 1.00 & 0.00 &
\xmark &
1.00 & 1.00 & 0.33 
& \xmark
\\ \hline
\multicolumn{1}{ |c ||}{\bf FOR} & 
0.00 & 0.50 & 0.50 &
\xmark &
1.00 & 0.33 & 1.00 
& \xmark
\\ \hline
\multicolumn{1}{ |c ||}{\bf AR} & 
0.50 & 0.50 & 0.50 &
\cmark &
0.50 & 0.25 & 0.75 
& \xmark
\\ \hline
\multicolumn{1}{ |c ||}{\bf Acc.} & 
1.00 & 0.25 & 0.75 &
\xmark &
0.00 & 0.50 & 0.50 
& \xmark
\\ \hline
\multicolumn{1}{ |c ||}{\bf Individual Fairness} & 
\multicolumn{4}{ c }{ \makecell{Rejects $\frac{2}{5}$ deserving users\\Accepts $\frac{2}{5}$ undeserving users} } & 
\multicolumn{4}{ |c |}{ \makecell{Rejects $\frac{3}{5}$ deserving users\\Accepts $\frac{3}{5}$ undeserving users} } 
\\ \hline
\end{tabular}

}
\vspace{-3mm}
 \caption{
			\small{[Top] A set of ten users along with their true labels, $y \in \{0,1\}$, and predicted labels, $\hat{y} \in \{0, 1\}$, by two classifiers $\mathcal{C}_1$ and $\mathcal{C}_2$.
			 				 	Label $1$ represents a \emph{more desirable} outcome (\eg, receiving a loan) than label $0$. 
			 				 	The users belong to three different groups: {\color{Red}${g_1}$ (red)}, {\color{Green} ${g_2}$ (green)}, and {\color{blue}${g_3}$ (blue)}.
			 				 								 										 				 										 			 	[Bottom] Fairness of the classifiers according to
		                various group- and individual-level metrics. The
		                group-level metrics are false positive rate (FPR),
		                false negative rate (FNR), false discovery rate (FDR),
		                false omission rate (FOR), acceptance rate in
		                desirable class (AR), accuracy (Acc.), while our
		                individual-level metric requires individuals deserving similar outcomes (i.e., with similar true lables) to receive similar outcomes (i.e., receive similar predicted labels).
The table also shows information
		                about whether the classifier is fair w.r.t. the
		                corresponding conditions or not. The fairness
		                conditions are described in detail in
		                Figure~\ref{fig:clf_fairness}.
		                We note that while both $\mathcal{C}_1$ and $\mathcal{C}_2$ are individually unfair
 according to our unfairness measure (with benefits defined as in 
		                		                Eq.~\ref{eq:adjust}
		                and Generalized Entropy in Eq.~\ref{eq:GE} used with $\alpha=2$), the unfairness of $\mathcal{C}_1$ is $0.2$ whereas the unfairness of $\mathcal{C}_2$ is $0.3$. Hence, $\mathcal{C}_2$ is more individually unfair than $\mathcal{C}_1$. One can similarly quantify and compare unfairness based on other fairness notions described in Figure~\ref{fig:clf_fairness}.
		                }
 }
 \label{fig:comparing_clfs}
\end{figure}

\begin{figure*}[ht]
\centering
\resizebox{0.8\textwidth}{!}{
\begin{tabular}{|c|c|c|c|c|c|c|c|}
\hline
& \multicolumn{2}{  c |}{{\bf Fairness}}  & {\bf Fairness} & \multicolumn{4}{c|}{\bf Benefit Function }  
\\ 
\cline{5-8}
& \multicolumn{2}{  c |}{ {\bf Notion} }  & {\bf Condition } & {\bf TP} & {\bf TN} & {\bf FP} & {\bf FN}
\\ 
\hline
\multirow{6}{*}{ \makecell{\bf Group\\ \bf Fairness} } & \multirow{5}{*}{ Parity Mistreatment}  & Accuracy & Equal accuracy for all groups 
& 1 & 1 & 0 & 0
\\ \cline{3-8}
&& \multirow{ 2}{*}{Equal Opportunity} & Equal FPR for all groups 
& n/a & 1 & 0 & n/a
\\ \cline{4-8}
&& & Equal FNR for all groups 
& 1 & n/a & n/a & 0
\\ \cline{3-8}
&& \multirow{ 2}{*}{Well-calibration} & Equal FDR for all groups
& 1 & n/a & 0 & n/a
\\ \cline{4-8}
&& & Equal FOR for all groups 
& n/a & 1 & n/a & 0
\\ \cline{2-8}
& \multicolumn{2}{c|}{Parity Impact / Statistical Parity} & Equal acceptance rate for all groups 
& 1 & 0 & 1 & 0
\\ \cline{1-8}
\makecell{{\bf Individual}\\ {\bf Fairness}}& \multicolumn{2}{c|}{\makecell{Our proposal}} & \makecell{Individuals \emph{deserving} similar \\ outcomes \emph{receive} similar outcomes} 
& 1 & 1 & 2 & 0
\\ \hline
\end{tabular}
}
\vspace{-2mm}
 \caption{
 	\small{
 	A summary of different fairness notions and their
    corresponding fairness conditions. We also show a benefit function
    that we use to compute inequality under each of the fairness
    conditions. Since all outcomes of a classifier can be decomposed into
    true positives (TP), true negatives (TN), false positives (FP) and
    false negatives (FN), the benefit function needs to assign a
    benefit score to each of these prediction types under any given
    fairness notion. For example, under statistical parity, which
    requires equal acceptance rates for all groups, we assign a benefit
    of ``$1$'' to TP and FP, and a benefit of ``$0$'' to TN and
    FN. ``n/a'' under an entry shows that these points are not
    considered under the corresponding fairness notion (\eg, equality
    of FPR requires considering only the points with negative true
    labels, \ie, $y=1$).
    }
 }
 \label{fig:clf_fairness}
\end{figure*}

\section{Measuring Algorithmic Unfairness\\ via Inequality Indices}\label{sec:axioms}
We first formally describe the setup of a fairness-aware machine learning task; then proceed to show that by defining an appropriate benefit function, existing inequality indices can be applied across the board to quantify algorithmic unfairness. We describe important properties (axioms) which we suggest a reasonable measure of algorithmic unfairness must satisfy. We end this section by comparing our proposed approach with previous work.

\subsection{Setting}
We consider the standard supervised learning setting: A learning algorithm receives a training data set $\Dcal=\{(\vx_i,y_i)\}_{i=1}^n$ consisting of $n$ instances, where $\vx_i \in \mathcal{X}$ specifies the feature vector\footnote{Throughout the paper, we use boldface notation to indicate a vector.} for an individual $i$ (\eg, $\vx_i$ could consist of individual $i$'s age, gender, and previous number of arrests in a criminal risk prediction task) and $y_i \in \mathcal{Y}$ is the outcome for this individual (\eg, whether or not they commit a bail violation). Unless specified otherwise, we assume $\mathcal{X} \subseteq \RR^M$, where $M$ denotes the number of features. 
If $\mathcal{Y}$ is a finite set of labels (\eg, $\mathcal{Y} =\{0,1\}$), the learning task is called classification; if $\mathcal{Y}$ is continuous (\ie, $\Ycal = \RR)$, it is called regression. In this paper, we will focus on \emph{binary classification}, but our work extends to multiclass classification and regression, as well.

We assume certain features (\eg, gender or race) are considered \emph{sensitive}.
Sensitive features specify an individual's membership in socially salient groups (\eg, women or African-Americans).
For simplicity of exposition, we assume there is just one sensitive feature. However, the discussion can be extended to account for multiple sensitive features.
We denote the sensitive feature for each individual $i$ as $z_i \in \Zcal = \{1, 2, \ldots, K\}$. 
Note that $z_i$ may or may not be part of the feature vector $\vx_i$.
One can define partitions of the dataset $\Dcal$ based on the sensitive feature, that is, $\Dcal_z = \{(\vx_i, y_i) \ |\  z_i = z\}$.
We refer to each partition $\Dcal_z$ of the data as a sensitive feature group.

The goal of a learning algorithm is to use the training data to fit a \emph{model} (or hypothesis) that accurately predicts the label for a new instance. A model $\theta: \mathcal{X} \rightarrow \mathcal{Y}$ receives the feature vector corresponding to a new individual and makes a prediction about his/her label. Let $\Thetab$ be the hypothesis class consisting of all the models from which the learning algorithm can choose.
A learning algorithm receives $\Dcal$ as the input; then utilizes the data to select a model $\theta \in \Thetab$ that minimizes some notion of loss. For instance, in classification the (0-1) loss of a model $\theta$ on the training data $\Dcal$ is defined as $L(\theta) = \sum_{i=1}^n \vert y_i - \hy_i \vert$
where $\hy_i = \theta(\vx_i)$.
The learning algorithm outputs $\theta^* \in \Thetab$ that minimizes the loss, \ie,
$\theta^* = \argmin L(\theta)$.

\subsection{Unfairness as Inequality in Benefits}
The core idea of our proposal is to quantify the unfairness of an
algorithm by measuring {how {\bf unequally} the outcomes of the
  algorithm {\bf benefit} different individuals or groups in a
  population}.
While intuitive, our proposal raises two key questions: (i) how should
we map algorithmic predictions received by individuals or groups to
{\bf benefits}? and (ii) given a set of benefits received by
individuals or groups, how should we quantify {\bf inequality} in the
benefit distribution?
We now tackle the first question, related to defining a benefit
function for an individual given an outcome.
In Section \ref{subsec:related_work}, we propose inequality indices as
the answer to the second question.

Our choice of the benefit function will be dictated by the type of
fairness notion we wish to apply on the task at hand. Figure \ref{fig:clf_fairness} 
summarizes the different fairness notions that have been defined in
prior works and their corresponding benefit functions. We now explain
the choice of our benefit functions for the different fairness notions
in the context of binary classification.
Formally, let $y_i \in \cY=\{0, 1\}$ indicate the true label for
individual $i$.
We assume that labels in the training data reflect \emph{ground
  truth}, and thus, $y_i$ is the label \emph{deserved} by individual
$i$.  Let $\hy_i \in \{0, 1\}$ be the label the algorithm assigns to
individual $i$.  

Intuitively, the algorithmic benefit an individual $i$ receives,
$b_i$, should capture the {\it desirability} of outcome $\hy_i$ for
the individual. The desirability of an individual's outcome may be
determined taking into account the individual's own preferences or the 
broader societal good.  For instance, consider the criminal risk
prediction example, where the positive label ($\hat{y}=1$) indicates a low
risk of criminal behavior and the negative label ($\hat{y}=0$) indicates a high
risk of criminal behavior. An individual defendant would clearly prefer
the former outcome over the latter. However, from a social good
perspective, accurate outcomes ($\hat{y}=y$) would be more desirable
than inaccurate outcomes. Furthermore, amongst the inaccurate
outcomes, one might wish to distinguish between the desirability of
false positives (where a high risk person is released) and false
negatives (where an low risk person is withheld).  

In our binary classification scenario, where all outcomes can be
decomposed into true positives ($\hat{y}=1, y=1$), true negatives
($\hat{y}=0, y=0$), false positives ($\hat{y}=1, y=0$), and false
negatives ($\hat{y}=0, y=1$), the choice of our benefit function
crucially determines the relative desirability of these different
types of outcomes and captures different notions of fairness. For
instance, the notion of {\it parity mistreatment} considers accurate
outcomes as more desirable than inaccurate ones -- so we choose a
benefit function that assigns higher value ($b_i=1$) to true positives
and true negatives and a lower value ($b_i=0$) to false positives and
false negatives. In contrast, the notion of {\it parity impact}
considers a positive label outcome as more desirable than a negative
label outcome -- so we adapt the benefit function to assign higher
value ($b_i=1$) to true positives and false positives and a lower
value ($b_i=0$) to true negatives and false negatives. To capture {\it
  group fairness}, once we define the benefits for all
individuals, $\veb=(b_1,\cdots,b_n)$, we can define the benefit for a
subset/group $g$ of the population, denoted by $\mu_g$, as the mean
value of the benefits received by individuals in the group: $\mu_g =
\frac{1}{|g|}\sum_{i \in g} b_i$.

To capture {\it individual fairness}, we propose defining
the benefit function of an individual $i$ as the {\it discrepancy}
between $i$'s preference for the outcome $i$ truly deserves (\ie,
$y_i$), and $i$'s preference for the outcome the learning algorithm
assigns (\ie, $\hat{y}_i$). As an illustration, in this work we consider
a benefit function that assigns the
highest value ($b_i=2$) for false positives (i.e., individuals that
receive the advantageous positive label undeservedly), moderate values
($b_i=1$) for true positives and true negatives (i.e., individuals
that receive the labels they deserve) and lowest value ($b_i=0$) for
false negatives (i.e., individuals that receive the disadvantageous
negative label despite deserving the positive label). More precisely, we compute the benefit for individual $i$ as follows:
\begin{equation}\label{eq:adjust}
b_i = \hy_i - y_i +1.
\end{equation}

We make two observations about the values of the benefit functions for
different types of outcomes. First, while different fairness notions
specify a preference ordering for different types of outcomes (i.e.,
true positives, false positives, true negatives, and false negatives),
the absolute benefit values could be specified differently. The choice
of benefit values would depend on the context and task at hand and the
difficulty of determining them may vary in practice.  Second, as many
existing measures of inequality in benefits are limited to handling
non-negative values, we need to ensure that $b_i \geq 0$ for
$i=1,\cdots, n$ and that there exists $j\in[n]$ such that $b_j>0$.

Our proposal is to measure the {\it overall individual-level}
unfairness of an algorithm by plugging $b_i$'s (as defined above) into
an existing inequality index (such as generalized entropy---to be
defined shortly). Throughout the rest of the paper, we will use the
terms ``overall unfairness'' and ``individual unfairness''
interchangeably, to refer to our proposed measure.  
Our approach can be further generalized to measuring (un)fairness
beyond supervised learning tasks (e.g. for unsupervised tasks, such as
clustering or ranking)---this only requires the specification a proper
notion of benefit for individuals given their \emph{relative} outcomes
within the population. We leave a careful exploration of this
direction for future, and focus on supervised learning tasks in the
current work.

Next, we discuss how we
can generally quantify the unfairness of an algorithm as the degree to
which it distributes benefit unequally across individuals using
inequality indices.

\subsection{Axioms for Measuring Inequality} \label{subsec:related_work}
Borrowing insights from the rich body of work on the axiomatic characterization of inequality indices in economics and social science~\citep{atkinson1970measurement,
sen1973economic, 
kolm1976unequal,
kolm1976unequali,
kakwani1980class,
cowell1981additivity,
litchfield1999inequality},
we argue that many axioms satisfied by inequality indices are appealing properties
for measures of algorithmic unfairness. Therefore, inequality indices are
naturally well-suited as measures for algorithmic unfairness. In this section, we briefly overview these axioms. 

Suppose society consists of $n$ individuals, where $b_i \geq 0$ denotes the benefit individual $i$ receives as the result of being subject to algorithmic decision making. An inequality measure,
$I: \bigcup_{n=1}^\infty \RR_{\geq 0}^n \rightarrow \RR_{\geq 0},$
maps any benefit distribution/vector $\veb$ to a non-negative real number $I(\veb)$.
A benefit vector $\veb$ is considered less unfair (\ie, more fair) than $\veb'$ if and only if $I(\veb) < I(\veb')$. 
Many inequality indices previously studied satisfy the following four principles:
\begin{itemize}
\item \textbf{Anonymity:} The measure does not depend on any characteristics of the individuals other than their benefit, and is independent of who earns each level of benefit. Formally:
$$I(b_1,b_2,\cdots,b_n) = I\left(b_{(1)},b_{(2)},\cdots,b_{(n)}\right),$$
where $(b_{(1)},b_{(2)},\dots,b_{(n)})$ is the benefit vector $(b_1,b_2,\cdots,b_n)$ sorted in ascending order.
\item \textbf{Population invariance:} The measure is independent of the size of the population under consideration. More precisely, let $\veb' = \langle \veb,\cdots,\veb \rangle \in \RR^{nk}_{\geq 0}$ be a $k$-replication of $\veb$. Then
$I(\veb) = I(\veb').$
\item \textbf{Transfer principle:} Transferring benefit from a high-benefit to a low-benefit individual must decrease inequality. More precisely for any $1\leq i<j \leq n$ and $0<\delta<\frac{b_{(j)} - b_{(i)}}{2}$, $$ I(b_{(1)},\cdots, b_{(i)} + \delta,\cdots,b_{(j)}-\delta, \cdots, b_{(n)})<I(\veb).$$
Note that the transfer should not reverse the relative position of the two
individuals $i$ and $j$. The transfer principle 
is sometimes called the Pigou-Dalton principle~ \citep{pigou1912wealth,dalton1920measurement}.
\item \textbf{Zero-Normalization:} The measure is minimized when every individual receives the same level of benefit. That is, for any $b \in \RR_{\geq 0}, $
$I(b,b,\cdots, b) = 0.$
\end{itemize}

\newcommand{\decompColWidth}{.7cm}

\begin{figure}[t]
\centering
\large
\renewcommand{\arraystretch}{.5}
\resizebox{\columnwidth}{!}{
\setlength\tabcolsep{1.5pt}
    \begin{tabular}{  | *2{>{\centering\arraybackslash\color{Red}}p{1.4cm} |} *4{>{\centering\arraybackslash\color{Green}}p{\decompColWidth} |} *4{>{\centering\arraybackslash\color{blue}}p{\decompColWidth} |}}
 	 	 	 							\hline
	$b_1$ & $b_2$  & $b_3$ & $b_4$ & $b_5$ & $b_6$ & $b_7$ & $b_8$ & $b_9$ & $b_{10}$
		\\ 
	\hline
	1 & 1 & 1 & 2 & 0 & 0 & 1 & 1 & 1 & 2
	\\ \hline
	\multicolumn{10}{  |c|  }{{\textbf{Overall individual-level unfairness}} = $I(b_1, \ldots, b_{10})$} \\ \hline
	\multicolumn{10}{  c  }{} \\ 
									\hline
	$\mu_{g_1}$ & $\mu_{g_1}$ & $\mu_{g_2}$ & $\mu_{g_2}$ & $\mu_{g_2}$ & $\mu_{g_2}$ & $\mu_{g_3}$ & $\mu_{g_3}$ & $\mu_{g_3}$ & $\mu_{g_3}$ 
	\\
	\hline
	1 & 1 & 0.75 & 0.75 & 0.75 & 0.75 & 1.25 & 1.25 & 1.25 & 1.25
		\\
	\hline
	\multicolumn{10}{  |c|  }{\makecell{{\textbf{Between-group unfairness}}  \\ = $I({\color{Red}\mu_{g_1}, \mu_{g_1}, }{\color{Green}\mu_{g_2}, \mu_{g_2}, \mu_{g_2}, \mu_{g_2}, }{\color{blue} \mu_{g_3}, \mu_{g_3}, \mu_{g_3}, \mu_{g_{3}}})$ }}
		\\ 
	\hline
	\multicolumn{10}{  c  }{} \\ 
							\hline
	$b_1$ & $b_2$  & $b_3$ & $b_4$ & $b_5$ & $b_6$ & $b_7$ & $b_8$ & $b_9$ & $b_{10}$
	\\
					\hline
	0 & 2  & 1 & 0 & 1 & 1 & 1 & 1 & 1 & 2 
	\\
	\hline
	\multicolumn{2}{  |c  }{  {\bf Within-group unf.}   } & 
	\multicolumn{4}{  |c  }{  {\bf Within-group unf.}   } & 
	\multicolumn{4}{  |c|  }{  {\bf Within-group unf.}   } 
	\\
	\multicolumn{2}{  |c  }{  =$I({\color{Red}b_1, b_2})$  } & 
	\multicolumn{4}{  |c  }{  =$I({\color{Green}b_3, b_4,b_5,b_6})$  } & 
	\multicolumn{4}{  |c|  }{ =$I({\color{blue}b_7, b_8,b_9,b_{10}})$  } 
	\\ 
	\hline	
		\multicolumn{10}{  c  }{} \\

								 \end{tabular}
 \setlength\tabcolsep{6pt}  }
 \vspace{-4mm}
 \caption{
 	\small{The set of ten users along with the benefit that each user receives from classifier $\mathcal{C}_1$ in Figure~\ref{fig:comparing_clfs}.
 	 	The overall individual-level unfairness of the classifier can be computed as the inequality $I$ over the benefits received by the users.
 	 	Overall unfairness can be decomposed into two components: 
 	 	1) between-group unfairness is computed as the inequality between (weighted) average group benefits for a given group,
 	 	and 2) within-group unfairness which is a weighted sum of within-group inequality. 
 	 	 	 	 		 		 		 		 		 		 		 		 		 		 		 		 		 	 }}
 \label{fig:unf_decomp}
 \end{figure}

In addition to the above four principles satisfied by many inequality indices, we also focus on the following property which is important for our purposes. \textbf{Subgroup decomposability} 
is a structural property of some inequality measures requiring that for any partition $G$ of the population into groups, the measure, $I(\veb)$, can be expressed as the sum of a \emph{``between-group component"} $I^G_\beta(\veb)$ (computed by assigning to each person in a subgroup $g \in G$ the subgroup's mean benefit $\mu_g$) and a \emph{``within-group component"} $I^G_\omega(\veb)$ (a weighted sum of subgroup inequality levels): \footnote{When the partition $G$ we are referring to is clear from the context, we drop the superscript $G$ to simplify notation.}
$$I(\veb)=I_\beta(\veb)+I_\omega(\veb).$$
See Figure~\ref{fig:unf_decomp} for an illustration of this property.

While not all inequality measures satisfy the decomposability property (\eg, the Gini Index does not),
the property has been studied extensively in economics, as it allows economists to compare patterns and dynamics of inequality in different subpopulations (\eg, racial minorities~\cite{conceiccao2000young}).

\xhdr{Our measure of unfairness}
For quantifying algorithmic unfairness, in this paper, we focus on a family of inequality indices called {\it  generalized entropy indices}. For a constant $\alpha\notin \{0,1\}$, the generalized entropy of benefits $b_1,b_2,\cdots,b_n$ with mean benefit $\mu$ is defined as follows:
\begin{equation}\label{eq:GE}
\Ecal^\alpha(b_1,b_2,\cdots,b_n) = \frac{1}{n\alpha(\alpha-1)} \sum_{i=1}^n \left[ \left(\frac{b_i}{\mu}\right)^\alpha -1\right].
\end{equation}
One can interpret generalized entropy as a measure of information theoretic \emph{redundancy} in data. 
Generalized entropy satisfies the earlier properties of anonymity, population-invariance, the Pigou-Dalton transfer principle, and zero-normalization. Further it is subgroup decomposable~\cite{cowell1981additivity}, and also \emph{scale-invariant}.\footnote{A measure $I$ is scale-invariant if for any constant $c>0$, $I(c\veb) = I(\veb)$.} 
In fact, \citet{shorrocks1980class} show that generalized entropy is the only differentiable family of inequality indices that satisfies population- and scale-invariance. Our interest in this family of inequality indices is motivated by this result and by our aim of understanding the trade-offs between individual and group-level unfairness.

\subsection{Comparison with Previous Work}\label{sec:rel_unfairness}
Existing notions of algorithmic fairness can be divided into two distinct categories: \emph{group} and \emph{individual} fairness.

\xhdr{Group fairness}
Group fairness notions require that given a classifier $\theta$, a certain group-conditional quality metric $q_z(\theta)$ is the same for all sensitive feature groups. That is:
$$q_z(\theta) = q_{z'}(\theta) \quad \forall z,z' \in \Zcal.$$
Different choices for $q_z(.)$  have led to different namings of the corresponding group fairness notions (see \eg, statistical parity~\cite{klein16,dwork2012fairness,Corbett-DaviesP17}, disparate impact~\cite{zafar_fairness,feldman_kdd15}, equality of opportunity~\cite{hardt_nips16}, calibration~\cite{klein16}, and disparate mistreatment~\cite{zafar_dmt}). Generally, these notions cannot guarantee fairness at the individual level, or when groups are further refined (see Kearns et al.~\cite{kearns2017preventing} for an illustrative example).

Existing group fairness notions are similar to the between-group component of fairness that we propose.
However, these notions usually do not take into account the size of different groups, whereas our between-group measure considers the proportion of the groups relative to the total population as illustrated in Figure~\ref{fig:unf_decomp}.
For example consider a population divided into two groups $A$ and $B$ containing 70\% and 30\% of the population with the negative ground truth label respectively.
Using generalized entropy with $\alpha=2$, a classifier $C_1$ achieving a false positive rate of $0.8$ on $A$ and $0.6$ on $B$ has a between-group inequality of $0.06$, whereas a classifier $C_2$ with false positive rates of $0.6$ on $A$ and $0.8$ on $B$ results in a lower between-group inequality of $0.04$.
However, when considering a group fairness measure based on differences in false positive rates between $A$ and $B$, $C_1$ and $C_2$ would be equally fair.

\xhdr{Individual fairness}
\citet{dwork2012fairness} first 
formalized the notion of individual fairness for classification tasks using Lipschitz conditions on the classifier outcomes. Their notion of individual fairness requires that two individuals who are similar with respect to the task at hand, receive similar classification outcomes. 
Dwork et al.'s definition is, therefore, formalized in terms of a similarity function between individuals. For instance, in practice given two individuals with feature values $\vx$ and $\vx'$, and suitable distance functions $D_\mathcal{X}$ and $D_\mathcal{Y}$ (defined over $\cX \times \cX$ and $\Delta(\cY) \times \Delta(\cY)$, respectively), Dwork et al.'s notion for individual fairness requires the following condition to hold:
\begin{align*}
D_\mathcal{Y}\big(p(\hat{y}=1 | \vx),p(\hat{y}=1 | \vx')\big) < D_\mathcal{X}(\vx, \vx'). \label{eq:sec2.dwork.ind}
\end{align*}
Due to its dependence on the individual feature vectors $\vx$, Dwork et al.'s notion of individual fairness does not satisfy the anonymity principle.

Furthermore,
Dwork et al.'s notion of individual fairness only provides a `yes/no' answer to whether fairness \emph{conditions} are satisfied, but does not provide a meaningful \emph{measure} of algorithmic fairness when considered \emph{independent of prediction accuracy}. We further illustrate this point with two examples:
First, by this definition a model that assigns the same outcome to everyone is considered fair, regardless of people's merit for different outcomes (e.g. awarding pretrial release to every defendant is considered fair, even though only some of them---those who appear for subsequent hearings and don't commit a crime\footnote{For making the pretrial release decisions, these two are the main criteria that the judges or the algorithms try to assess~\cite{summers2010pretrial,berk2017fairness}.}---deserve to be awarded the pretrial release).
Second, the definition does not take into account the difference in social desirability of various outcomes. For instance, if one flips the (binary) labels predicted by a fair classifier, the resulting classifier will be considered equally fair (e.g. a classifier that awards pretrial release to a defendant if and only if they go on to violate the release criteria is considered fair!).
The measure we propose in equations~\ref{eq:adjust} and \ref{eq:GE} addresses these issues by offering a merit-based metric of fairness that seeks to \emph{equalize} the \emph{benefit} individuals receive as the result of being subject to algorithmic decision making. 
Finally, we remark that there has been interest in a similar axiomatic approach to methods for algorithmic interpretability \cite{Sun17,lundberg2017unified}.

\section{Theoretical Characterization}\label{sec:theoratical}

In this section, we characterize the conditions under which there is a tradeoff between accuracy and our notion of algorithmic fairness. Further, we shed light on the relationship between our notion of fairness and existing group measures, precisely connecting the two when the inequality index in use is
\emph{additively decomposable} (see~\cite{subramanian2011inequality} and the references therein). At a high level, we show that group unfairness is one piece of a larger puzzle: overall unfairness may be regarded as a combination of unfairness \emph{within-} and \emph{between}-groups. As the number of groupings increases, with each becoming smaller (eventually becoming single individuals), the between-group component grows to be an increasingly large part of the overall unfairness.  

\subsection{Accuracy vs. Individual Fairness}
We begin by observing that the fairness optimal classifier is perfectly fair if and only if the accuracy optimal classifier is perfectly accurate. 
Given a classifier $\theta$ and training data set $\Dcal=\{(\vx_i,y_i)\}_{i=1}^n$, let $I_\Dcal(\theta)$ specify the individual unfairness of $\theta$ on $\Dcal$, that is, $I_\Dcal(\theta) = I(b^\theta_1,\cdots,b^\theta_n)$ where $b^\theta_i = 1 + \theta(\vx_i) - y_i$.
$L_\Dcal(\theta)$ is the empirical loss of $\theta$ on $\Dcal$.
\begin{proposition}\label{prop:bayes}
Suppose $I(.)$ is a zero-normalized inequality index and $\Thetab$ is closed under complements.\footnote{In the context of a binary classification task with $\cY = \{0,1\}$, $\Thetab$ is closed under complements if for any $\theta \in \Thetab$, also $1-\theta \in \Thetab$.} For any training data set $\Dcal$, there exists a classifier $\theta \in \Thetab$ for which $I_\Dcal(\theta)=0$ if and only if there exists a classifier $\theta'$ for which $L_\Dcal(\theta')=0$.
\shortLongVersion{\footnote{Proofs are omitted due to space constraints, and can be found in the long version of the paper available online on arXiv.}}{\footnote{Proofs can be found in the appendix.}}
\end{proposition}
Proposition \ref {prop:bayes} may seem to suggest that our notion of fairness is entirely in harmony with prediction accuracy: by simply minimizing prediction error, unfairness will be automatically eliminated. While this is true in the special case of fully separable data (or when we have access to an oracle with 0 prediction error), it is not true in general. The following result shows that under broad conditions, the fairness optimal classifier may not coincide with the accuracy optimal classifier. 
For training data set $\Dcal$, 
let $\theta^A_\Dcal $ be the accuracy optimal classifier, and $\theta^F_\Dcal$ be the fairness optimal classifier:
$$\theta^A_\Dcal = \arg \min_{\theta} L_\Dcal(\theta) \text{ and } \theta^F_\Dcal = \arg \min_{\theta} I_\Dcal(\theta).$$
\begin{proposition}\label{prop:bayes2}
Suppose $I(.)$ satisfies the transfer principle and is population- and scale-invariant. 
If there exists a feature vector $\tilde{\vx}$ such that
$0 < \Prob[y =1 \vert \vx=\tilde{\vx} ] < 0.5$, then there exists a training data set $\Dcal$ for which $\theta^B_\Dcal \not\equiv \theta^F_\Dcal$.
\end{proposition}
Even though the fairness optimal and accuracy optimal classifiers do not necessarily coincide, one might wonder if the fairness optimal classifier always results in near-optimal accuracy. In fact, it does not. \shortLongVersion{An example in the long version of the paper}{Example \ref{ex:fair_accuracy} in the appendix} shows that the accuracy of the fairness optimal classifier can be arbitrarily worse than that of the accuracy optimal classifier.

\subsection{Individual vs. Group Fairness}
Next, we focus on additive-decomposability and show how this property allows us to establish formally the existence of tradeoffs between individual- and group-level (un)fairness. Suppose we partition the population into $|G|$ disjoint subgroups, where subgroup $g \in G$ consists of $n_g$ individuals with the benefit vector $\veb^g = (b^g_1,\cdots,b^g_{n_g})$ and mean benefit $\mu_g$. Each partition could, for instance, correspond to a sensitive feature group (\emph{e.g.}, $g=1$ consists of all African-American defendants and $g=2$, all white defendants). One can re-write the Generalized Entropy as follows:
{\small
\begin{eqnarray*}
\Ecal^\alpha(b_1,b_2,\cdots, b_n) &=&\sum_{g=1}^{|G|} \frac{n_g}{n}\left(\frac{\mu_g}{\mu}\right)^\alpha \Ecal^\alpha(\veb^g) \label{eq:GE_within}\\
&& + \sum_{g=1}^{|G|} \frac{n_g }{n \alpha(\alpha-1)}\left[ \left(\frac{\mu_g}{\mu}\right)^\alpha - 1\right] \label{eq:GE_between}\\ 
&=& \Ecal^\alpha_\omega(\veb)+ \Ecal^\alpha_\beta(\veb). \nonumber
\end{eqnarray*}
}
Note that imposing a constraint on a decomposable inequality measure, such as $\Ecal^\alpha(\veb)$, guarantees both within-group and between-group inequality are bounded. Existing notions of group fairness, however, capture only the \emph{between-group} component (when $|G|=2$, the between-group unfairness is minimized if and only if the two groups receive the same treatment on average). The problem with imposing a constraint on the between-group component ($\Ecal^\alpha_\beta(\veb)$) alone, is that it may drive up the within-group component, $\Ecal^\alpha_\omega(\veb)$. In fact, we show that if an individual-fairness optimal classifier is not group-fairness optimal, then optimizing for group fairness alone will certainly increase unfairness within groups sufficiently so as to raise the overall (individual) unfairness.

Formally, minimizing our notion of individual unfairness while guaranteeing a certain level of accuracy corresponds to the following optimization:\footnote{For simplicity, we are dropping the training data set $\Dcal$ from the subscripts.}
\begin{equation}\label{eq:ind}
\min_{\theta \in \Thetab}  \quad I_\beta(\theta)+I_\omega(\theta) \qquad
 \text{s.t. } \quad L(\theta) \geq \delta, \phantom{.}
\end{equation}
while minimizing only between-group unfairness corresponds to:
\begin{equation}\label{eq:gr}
\min_{\theta \in \Thetab} \quad I_\beta(\theta) \phantom{+I_\omega(\theta)}  \qquad
 \text{s.t. } \quad  L(\theta) \geq \delta. \phantom{,}
\end{equation}
 Let $\theta^*(\delta)$ be an optimal solution for optimization (\ref{eq:ind})---if there are multiple optimal solutions, pick one with the lowest $I_\beta$. Let $\theta^*_\beta(\delta)$ be any optimal solution for optimization (\ref{eq:gr}). The following holds.
\begin{proposition}\label{prop:group}
Suppose $I(.)$ is additively decomposable. 
For any $\delta \in [0,1]$, if $I_\beta\left(\theta^*_\beta(\delta)\right)\neq I_\beta\left(\theta^*(\delta)\right)$, then $I_\omega\left(\theta^*_\beta(\delta)\right)> I_\omega\left(\theta^*(\delta)\right)$ and $I\left(\theta^*_\beta(\delta)\right)> I\left(\theta^*(\delta)\right)$.
\end{proposition}

Next, we show that the contribution of the between-group component to overall inequality, i.e. $I_\beta / {I}$, depends---among other things---on the granularity of the groups. In particular, the between-group contribution increases with intersectionality: $I_\beta$ is lower when computed over just race (African-Americans vs. Caucasians), and is higher when computed over the intersection of race and gender (female African-Americans, \dots,  male Caucasians).
More precisely, suppose $G, G'$ are two partitions of the population into disjoint groups. Let $G\times G'$ specify the Cartesian product of the two partitions: for $g \in G, g' \in G'$, $i \in (g,g')$ if and only if $i \in g$ and $i \in g'$. It is easy to show the following result.
\begin{proposition}\label{prop:intersection}
Suppose $I(.)$ is zero-normalized and additively decomposable. 
Suppose $G, G'$ are two different partitions of the population into disjoint groups. For any benefit distribution $\veb$, 
$I^G_\beta(\veb) \leq I^{G \times G'}_\beta(\veb).$
\end{proposition}
 If one continues refining the groups, eventually every individual will be in their own group and the between-group unfairness becomes equivalent to the overall individual unfairness. This offers a framework to interpolate between group and individual fairness. 
When the number of groups is small and people within each group receive highly unequal benefits, the contribution of the between-group component to overall unfairness is small, and narrowing down attention to reducing $I_\beta$ alone may result in fairness gerrymandering~\cite{kearns2017preventing}: 
while it is often easy to reduce group unfairness in this case, doing so will affect the overall unfairness in unpredictable ways---potentially making the within-group unfairness worse. 
On the other hand, as the number of groups increases or the treatment of people within a group becomes more uniform, the role that the between-group component plays in overall unfairness grows -- but, as noted by Kearns et al.~\cite{kearns2017preventing}, it also becomes   computationally harder to control and limit the between-group unfairness. 

\begin{proposition}\label{prop:group_share}
Suppose $I(.)$ is zero-normalized and additively decomposable. 
Suppose $I(\veb) \neq 0$. For any partition $G$ of the population to disjoint subgroups, $0 \leq \frac{I^G_\beta(\veb)}{I(\veb)} \leq 1$. Further, there exist benefit distributions $\veb$ and $\veb'$ such that $\frac{I^G_\beta(\veb)}{I(\veb)} = 1$ and $\frac{I^G_\beta(\veb')}{I(\veb')} = 0$.
\end{proposition}

\paragraph{Implication for practitioners: individual or group unfairness?} We saw that the contribution of the group component to  overall unfairness is a nuanced function of the granularity with which the groups are defined, as well as the unfairness within each group. If our goal is to reduce overall unfairness, note that existing fair learning models that exclusively focus on reducing between-group unfairness would help only when between-group accounts for a large part of the overall unfairness.
Our measures of unfairness present a framework to examine this condition.

\section{Empirical Analysis}\label{sec:empirical}

In this section, we empirically validate our theoretical propositions from Section~\ref{sec:theoratical} on multiple real-world datasets.
Specifically, in Section~\ref{sec:empirical:fairness_acc_tradeoff}, our goal is to shed light on tradeoffs between overall individual-level unfairness and accuracy. In Section~\ref{sec:empirical:decomposability}, we explore how the overall unfairness decomposes along the lines of sensitive attribute groups. We use the subgroup-decomposability of our proposed unfairness measures to study finer-grained fairness-accuracy tradeoffs at the levels of between-group and within-group unfairness. In Section~\ref{sec:empirical:interaction}, we empirically explore how methods to control between-group unfairness affect other unfairness components.

\begin{figure*}[ht]
    \centering
    \begin{subfigure}[b]{\textwidth}
        \centering
        \begin{subfigure}[b]{0.36\textwidth}
            \includegraphics[width=\textwidth]{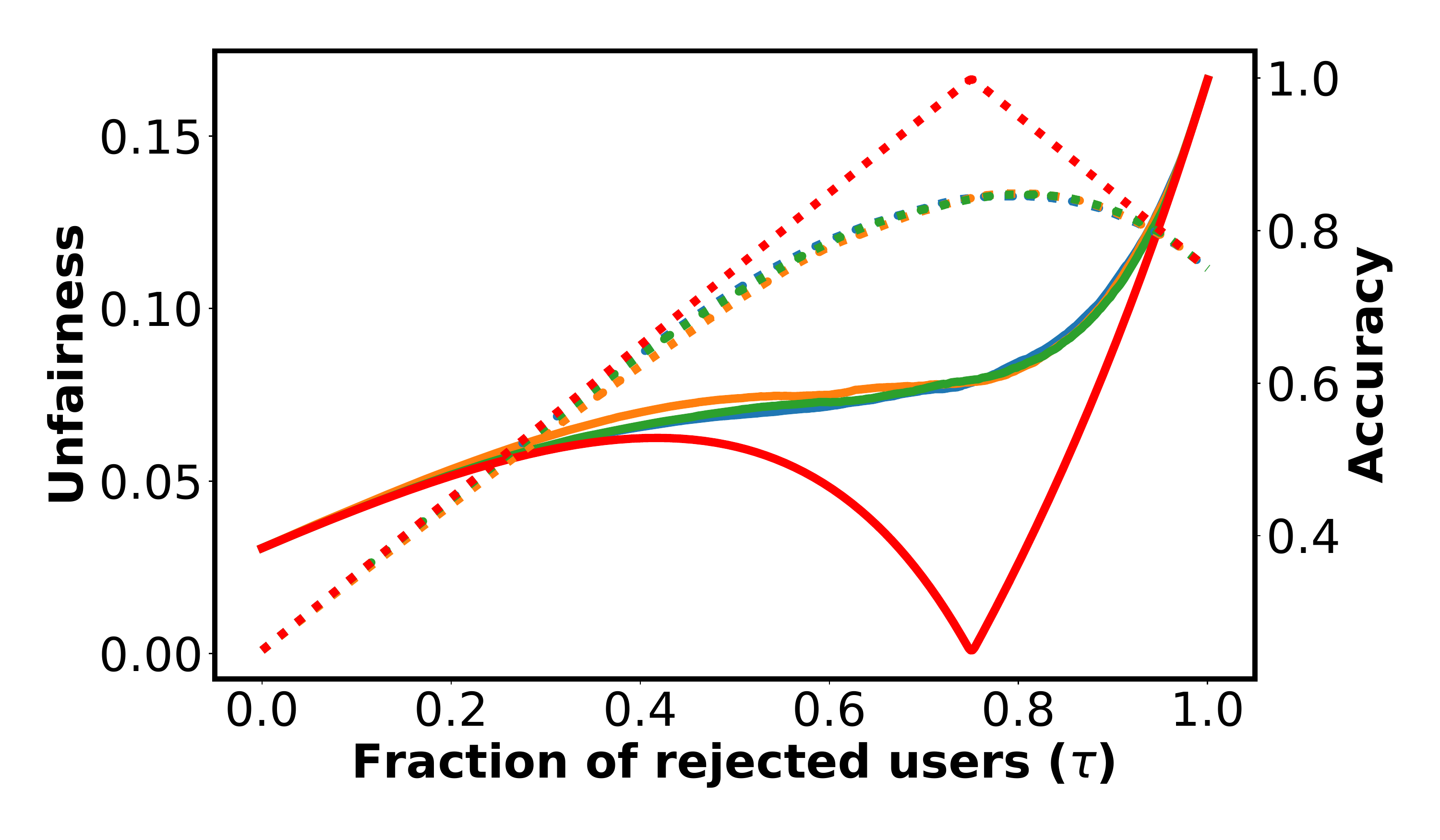}
        \end{subfigure}
        \begin{subfigure}[b]{0.36\textwidth}
            \includegraphics[width=\textwidth]{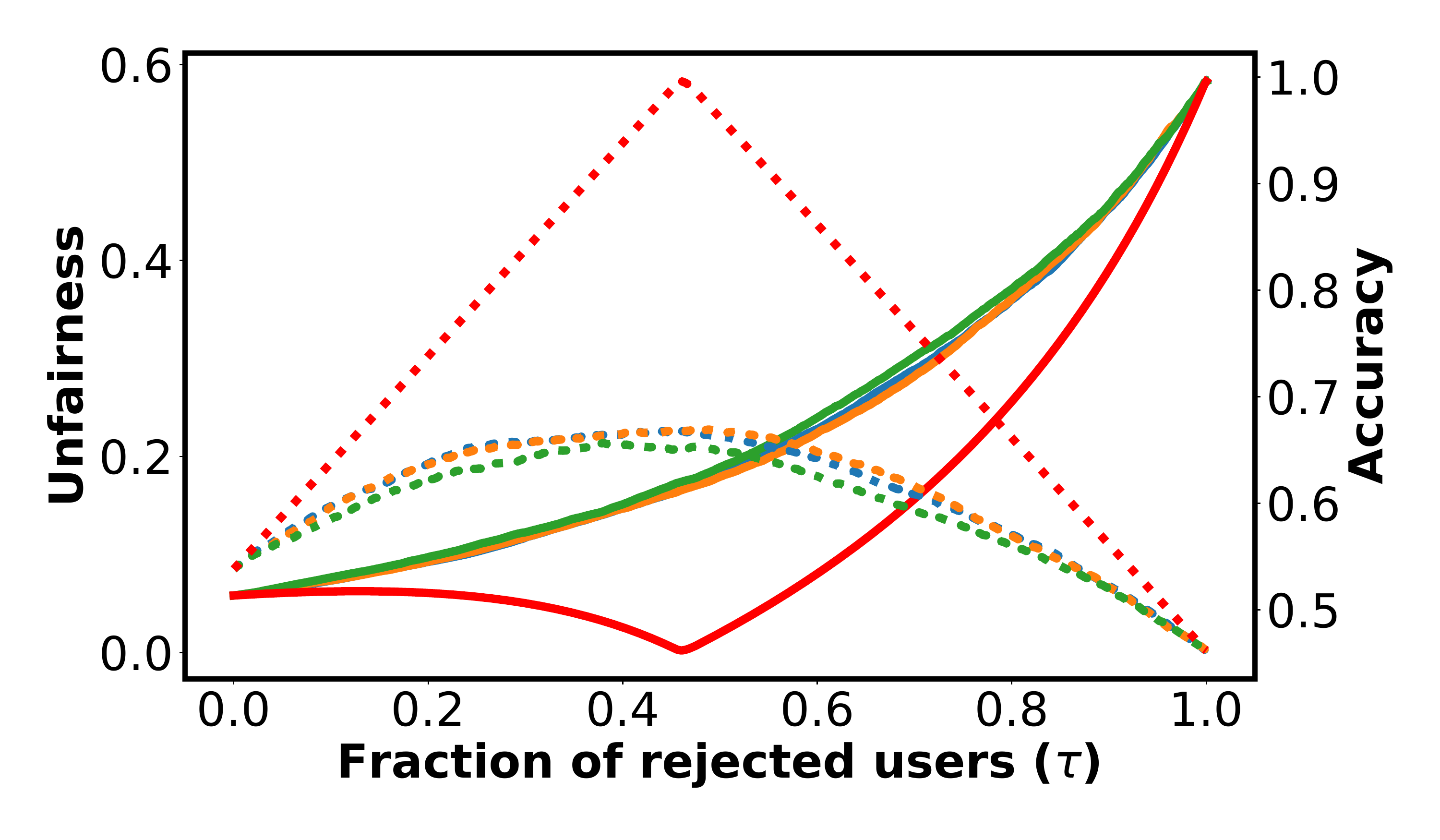}
        \end{subfigure}
    \end{subfigure}
    \begin{subfigure}[b]{\textwidth}
        \centering
        \begin{subfigure}[b]{0.37\textwidth}
            \includegraphics[width=\textwidth]{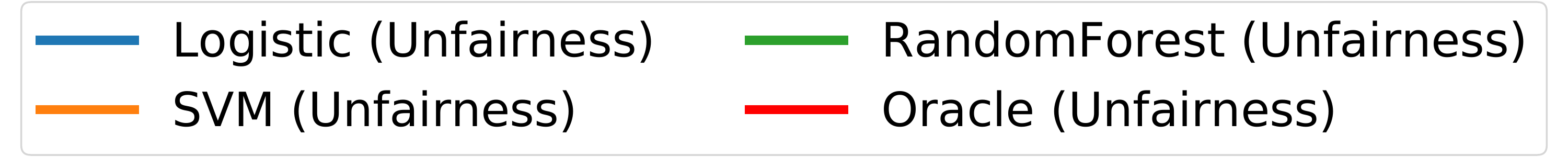}
            \caption{Adult dataset}
            \end{subfigure}
        \begin{subfigure}[b]{0.37\textwidth}
            \includegraphics[width=\textwidth]{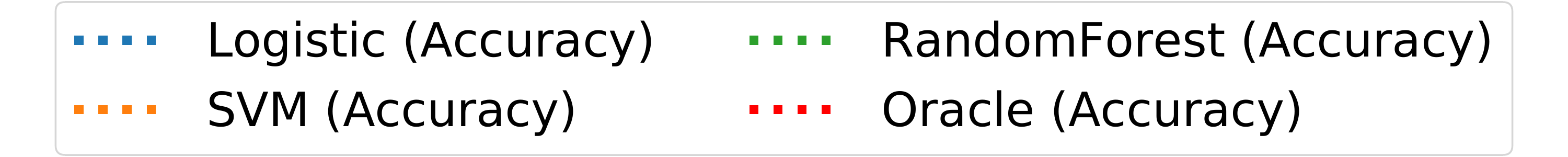}
            \caption{COMPAS dataset}
            \end{subfigure}
    \end{subfigure}
     \vspace{-7mm}
    \caption{
        \small{
        Overall unfairness (solid lines---$\Ecal^2(\veb)$  in Eq.~\ref{eq:empirical:gr2}) and 
        accuracy (dotted lines)  as a function of the decision ranking threshold ($\mathbf{\tau}$) for various classifiers. The positive and negative class ratio in the Adult dataset is about $0.25:0.75$, hence the $1.00$ accuracy point corresponds to $\tau=0.75$. Similarly, the positive and negative class ratio in the COMPAS dataset is about $0.55:0.45$, hence the $1.00$ accuracy point corresponds to $\tau=0.45$. For oracle, the optimal point for accuracy corresponds to minimal unfairness; this doesn't hold for other classifiers.     
                }
            }\label{fig:indiv_accuracy}
\end{figure*}

\xhdr{Setup and Datasets} We use the Generalized Entropy index (cf. Eq.~\ref{eq:GE}) with $\alpha=2$ (in other words, half the squared coefficient of variation) to measure unfairness:
\begin{equation}\Ecal^{2}(b_1,b_2,\cdots,b_n) = \frac{1}{2\times n} \sum_{i=1}^n \left[ \left(\frac{b_i}{\mu}\right)^2 -1\right].\nonumber
\end{equation}
As noted in Section~\ref{sec:axioms}, the Generalized Entropy index can be further decomposed into between-group and within-group unfairness as:
\begin{eqnarray}
\Ecal^2(b_1,b_2,\cdots, b_n) =& \Ecal^2_\omega(\veb)+ \Ecal^2_\beta(\veb). \label{eq:empirical:gr2}
\end{eqnarray}
We will refer to the quantity $\Ecal^2$ as \emph{individual unfairness} or \emph{overall unfairness} interchangeably, $\Ecal^2_\beta$ as between-group unfairness, and $\Ecal^2_\omega$ as within-group unfairness.

We experiment with two real-world datasets: 
(i) the {\it Adult income} dataset~\cite{adult_dataset}, and (ii) the {\it ProPublica COMPAS} dataset~\cite{propublica_data}. Both datasets have received previous attention~\cite{zafar_fairness,zafar_dmt,Corbett-DaviesP17,icml2013_zemel13,feldman_kdd15}. 

For the {\it Adult income} dataset, the task is to predict whether an individual earns more (positive class) or less (negative class) than 50,000 USD per year based on features like education level and occupation. We consider gender (female and male) and race (Black, White and Asian) as sensitive features. We filter out races (American Indian and Other) which constitute less than $1\%$ of the dataset. After the filtering, the dataset consists of $44,434$ subjects and $11$ features.

For the {\it ProPublica COMPAS} dataset, the task is to predict whether (negative class) or not (positive class) a criminal defendant would commit a crime within two years based on features like current charge degree or number of prior offenses. We use the same set of features as Zafar et al.~\cite{zafar_dmt}. The sensitive features in this case are also gender (female and male) and race (Black, Hispanic, White). The dataset consists of $5,786$ subjects and $5$ features.

For all experiments, we repeatedly split the data into $70\%$-$30\%$ train-test sets 10 times and report average statistics. All hyperparameters are validated using a further $70\%$-$30\%$ split of the train set into train and validation sets.

\begin{figure*}[ht]
    \centering
    \begin{subfigure}[b]{\textwidth}
        \centering
        \begin{subfigure}[b]{0.38\textwidth}
            \includegraphics[width=\textwidth]{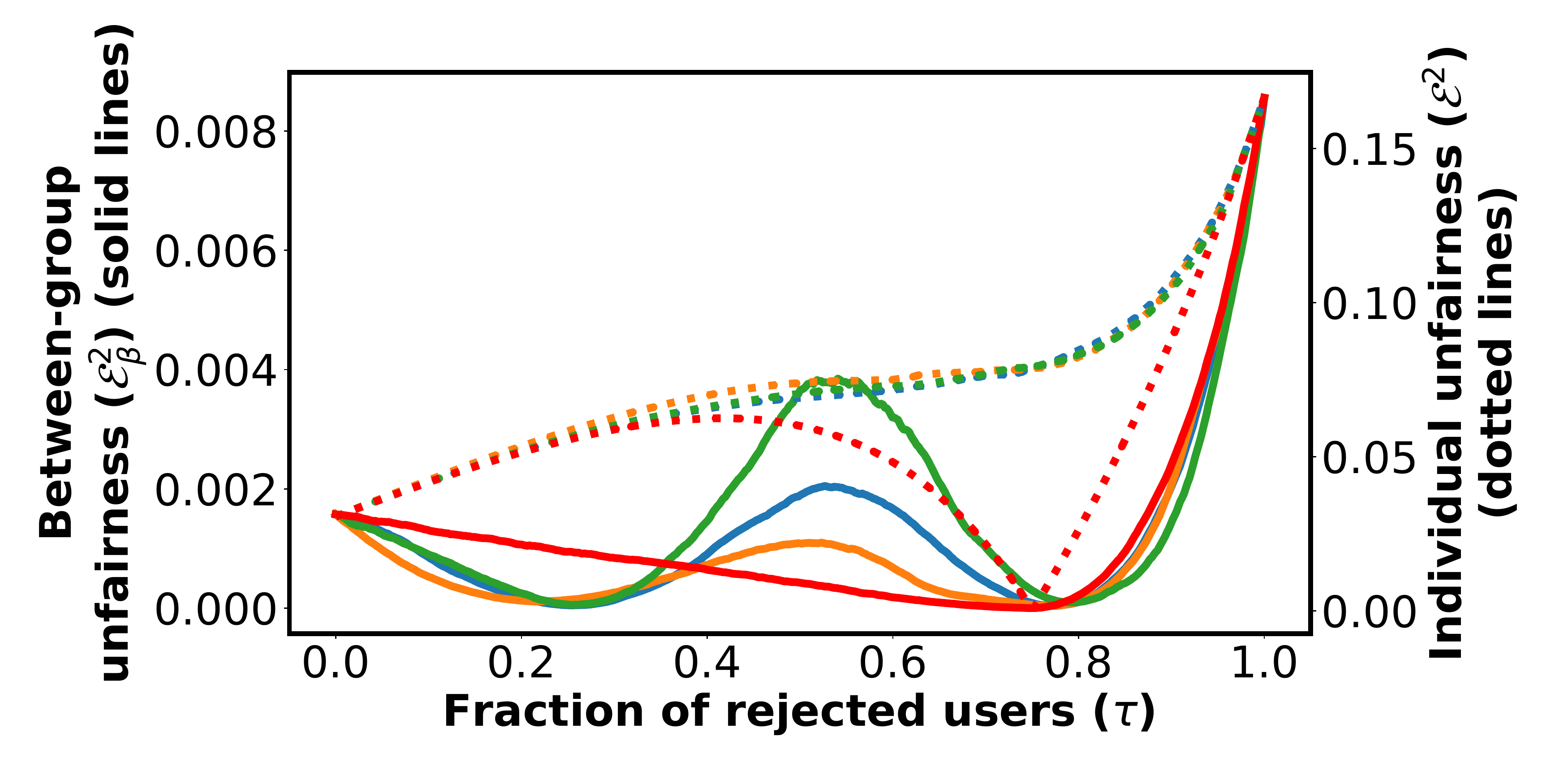}
        \end{subfigure}
        \begin{subfigure}[b]{0.38\textwidth}
            \includegraphics[width=\textwidth]{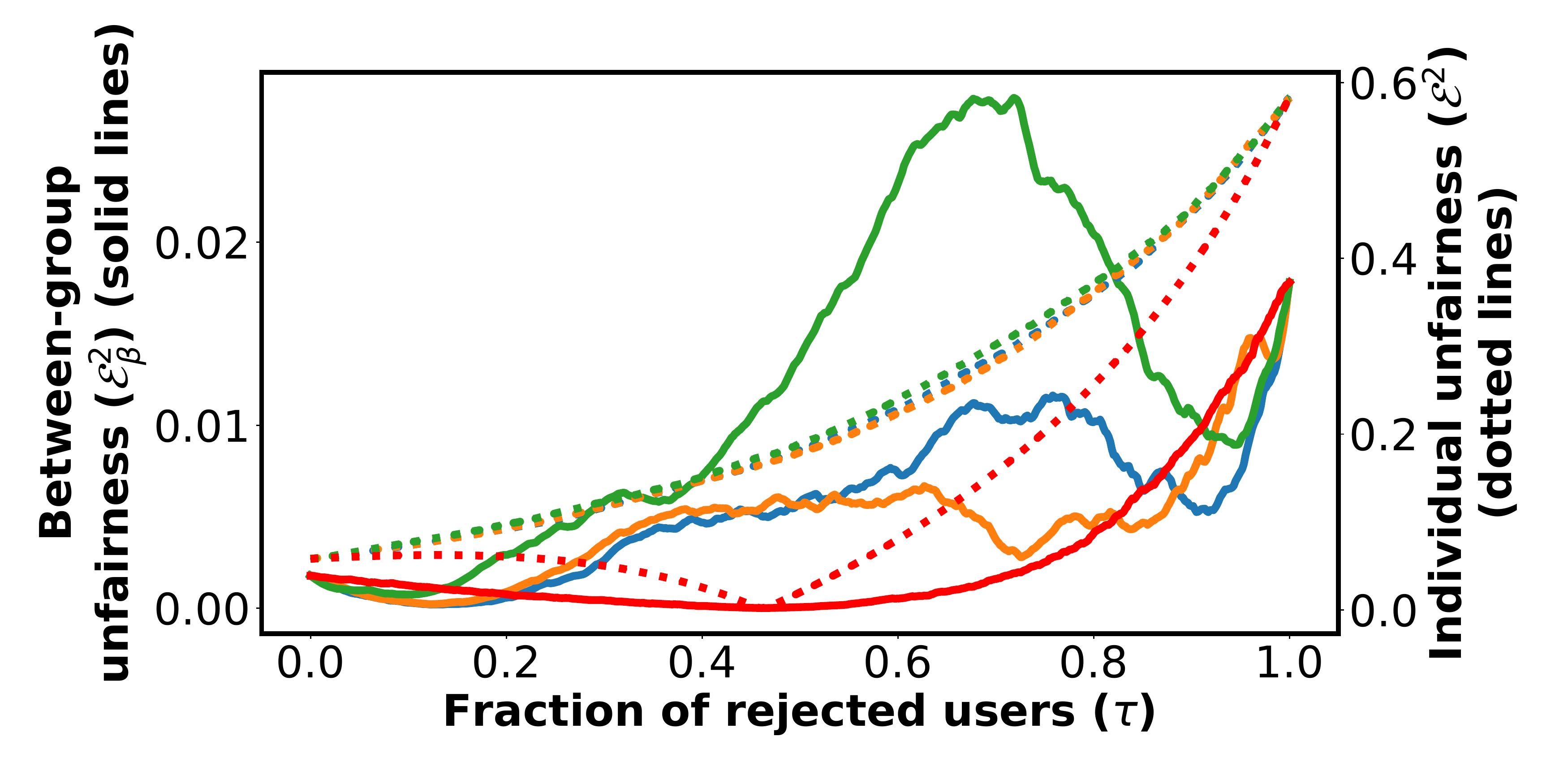}
        \end{subfigure}
    \end{subfigure}
    \begin{subfigure}[b]{\textwidth}
        \centering
        \begin{subfigure}[b]{0.38\textwidth}
            \includegraphics[width=\textwidth]{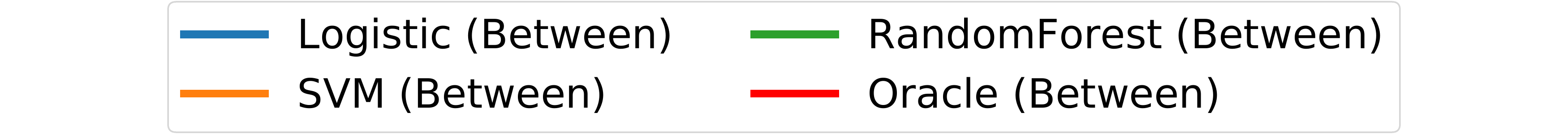}
            \caption{Adult dataset}
                                \end{subfigure}
        \begin{subfigure}[b]{0.38\textwidth}
            \includegraphics[width=\textwidth]{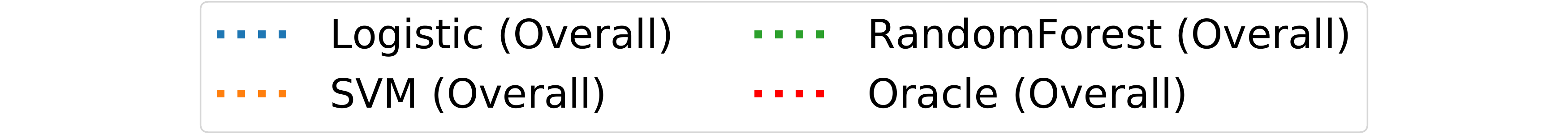}
            \caption{COMPAS dataset}
                                \end{subfigure}
    \end{subfigure}
     \vspace{-7mm}
    \caption{
    \small{Between-group unfairness (solid lines---$\Ecal^2_\beta(\veb)$  in Eq.~\ref{eq:empirical:gr2}) and overall unfairness (dotted lines---$\Ecal^2(\veb)$  in Eq.~\ref{eq:empirical:gr2}) as a function of the decision ranking threshold ($\mathbf{\tau}$) for various classifiers.
            Between-group unfairness $\Ecal^2_\beta(\veb)$ only constitutes a small fraction of the overall unfairness $\Ecal^2(\veb)$.}
    }
    \label{fig:within-group-accuracy}
    \vspace{-4mm}
\end{figure*}

\subsection{Fairness vs. Accuracy Tradeoffs} \label{sec:empirical:fairness_acc_tradeoff}
We begin by studying the tradeoff between the accuracy and the overall \textbf{\textit{individual unfairness}} of a given classifier ($\Ecal^2(\veb)$  in Eq.~\ref{eq:empirical:gr2}).
We use three standard classifier models: logistic regression, support vector machine with RBF kernel (SVM), and random forest classifier. Results in Section~\ref{sec:empirical:fairness_acc_tradeoff} and Section~\ref{sec:empirical:decomposability} are computed by optimizing these classifiers for accuracy.

Each of the above models computes the \emph{likelihood} of belonging to the positive class for every instance. We denote this likelihood by $p_i$ for an individual $i$.
We compare the fairness and accuracy of these classifiers with that of
an 
``oracle''
that can perfectly predict the label for every instance (and assigns $p_i \in \{0,1\}$). To predict a label in $\{0,1\}$ for individuals, we first rank all instances (in increasing order) according to their $p_i$ values with ties broken randomly; then we designate a decision ranking threshold $0 \leq \tau \leq 1$ and output a label of $1$ for individual $i$ 
if and only if $rank(i) \geq n\tau$, where $rank(i)$ denotes the rank of individual $i$ in the sorted list. 
In other words, an increasing value of $\tau$ corresponds to the classifier rejecting more people from the sorted list (in order of their positive class likelihood $p_i$).
As we vary $\tau$, we expect both accuracy as well as unfairness of the resulting predictions to change as discussed below and shown in Figure~\ref{fig:indiv_accuracy}.

For the oracle, as expected from Proposition~\ref{prop:bayes}, 
\emph{a perfect accuracy corresponds to zero unfairness}:
with an increasing $\tau$, the accuracy increases while the unfairness decreases. After a certain optimal value of threshold  $\tau$ (close to $0.75$ in the Adult data and $0.45$ in the COMPAS data), the trend reverses. We note that $0.75$ and $0.45$ represent the fraction of instances in the negative class in the respective datasets. Hence, at these optimal thresholds, all of the oracle's predictions are accurate (since the points are ranked based on their positive class likelihood) resulting in $0$ unfairness.

However, for all other (non-oracle) classifiers, as expected via Proposition~\ref{prop:bayes2}, the trend is very different: \emph{the optimal threshold for (imperfect) accuracy  is far from the optimal threshold for unfairness}. 
Moreover, with increasing $\tau$, while unfairness continually increases, for accuracy we initially see an increase followed by a drop.  We note that the overall unfairness is not always a monotone function of the decision ranking threshold \shortLongVersion{(an example is provided in the long version of the paper)}{as illustrated in Example~\ref{ex:fair_accuracy}}.

\subsection{Fairness Decomposability}\label{sec:empirical:decomposability}
As Figure~\ref{fig:unf_decomp} and Eq.~\ref{eq:empirical:gr2} show, the individual unfairness of a predictor can be decomposed into between-group and within-group unfairness. In this part, we study the \textit{\textbf{between-group unfairness}} component ($\Ecal^2_\beta(\veb)$  in Eq.~\ref{eq:empirical:gr2}) as we change $\tau$. To this end, we consider two sensitive features: gender and race. We split each of the datasets into all possible disjoint groups based on these sensitive features (\eg, White women, Hispanic men, Black women).

Figure~\ref{fig:within-group-accuracy} shows between-group unfairness along with overall unfairness for different values of $\tau$.
We notice that for the Adult dataset, the \emph{between-group unfairness follows a multi-modal trend}:  it starts from a non-zero value at $\tau=0$, falls to almost $0$ for most classifiers (except for the oracle) at around $\tau=0.2$, reaches a local peak again around $\tau=0.5$, and finally completes another cycle to fall and then reach its maximum value at $\tau=1.0$. The COMPAS dataset also shows a similar trend, albeit to a lesser extent.

\xhdr{Between-group unfairness and overall unfairness}
Comparing  the between-group unfairness and overall unfairness in Figure~\ref{fig:within-group-accuracy} reveals a very interesting insight:
for the same value of $\tau$,
the between-group unfairness (solid lines) is a very small fraction of the overall unfairness (dotted lines). For example, considering the performance of the logistic regression classifier on the COMPAS dataset, the maximum value of the \emph{overall unfairness} is close to $0.6$ whereas the maximum value of the \emph{between-group unfairness} is merely $0.01$. We hypothesize that since the number of sensitive feature-based groups is much smaller than the number of all individuals in the dataset, the individual unfairness value dominates the between-group unfairness.

\begin{figure}[hb]
    \centering
    \includegraphics[width=0.9\columnwidth]{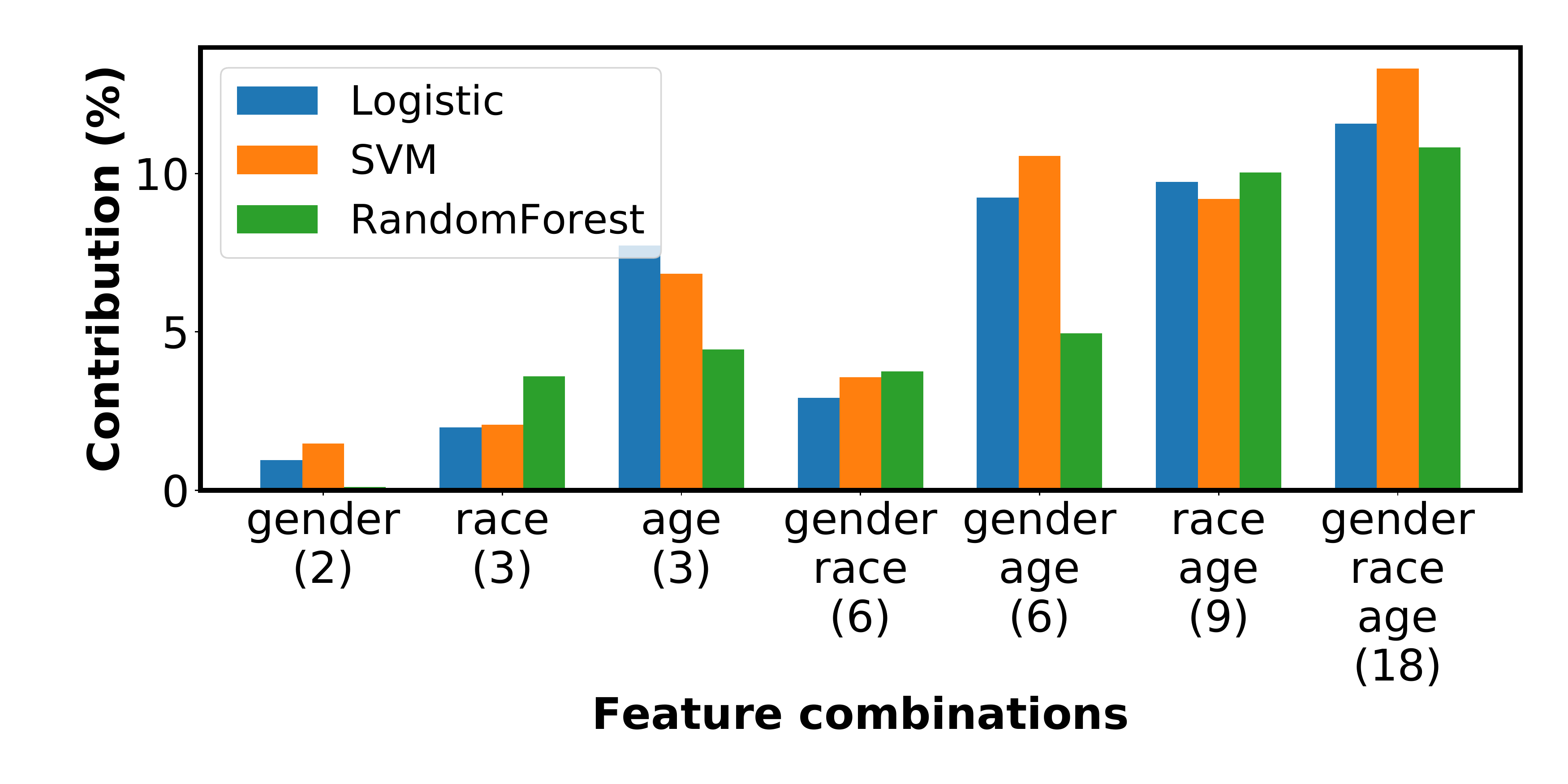}
     \vspace{-5mm}
    \caption{\small{Between-group unfairness ($\Ecal^2_\beta(\veb)$  in Eq.~\ref{eq:empirical:gr2}) as a fraction of the overall/individual unfairness ($\Ecal^2(\veb)$  in Eq.~\ref{eq:empirical:gr2}) for various combinations of sensitive features. 
    Numbers on the x-axis denote  how many sensitive feature groups would be formed when using the corresponding sensitive feature set. Logistic regression, SVM and Random Forests achieve similar accuracies of 66\%, 67\% and 65\% as well as similar overall individual unfairness of 0.145, 0.151 and 0.134 respectively on the ProPublica Compas dataset.
    }}
    \label{fig:frac_btwn_grp}
    \vspace{-5mm}
\end{figure}

\begin{figure*}[ht]
        \begin{subfigure}[b]{\textwidth}
    \centering
                            		                                                      	                                                       \centering
            \begin{subfigure}[b]{.30\textwidth}
            	\centering
	            \hspace{14mm}
                \textbf{Between-group}
                \includegraphics[width=\textwidth]{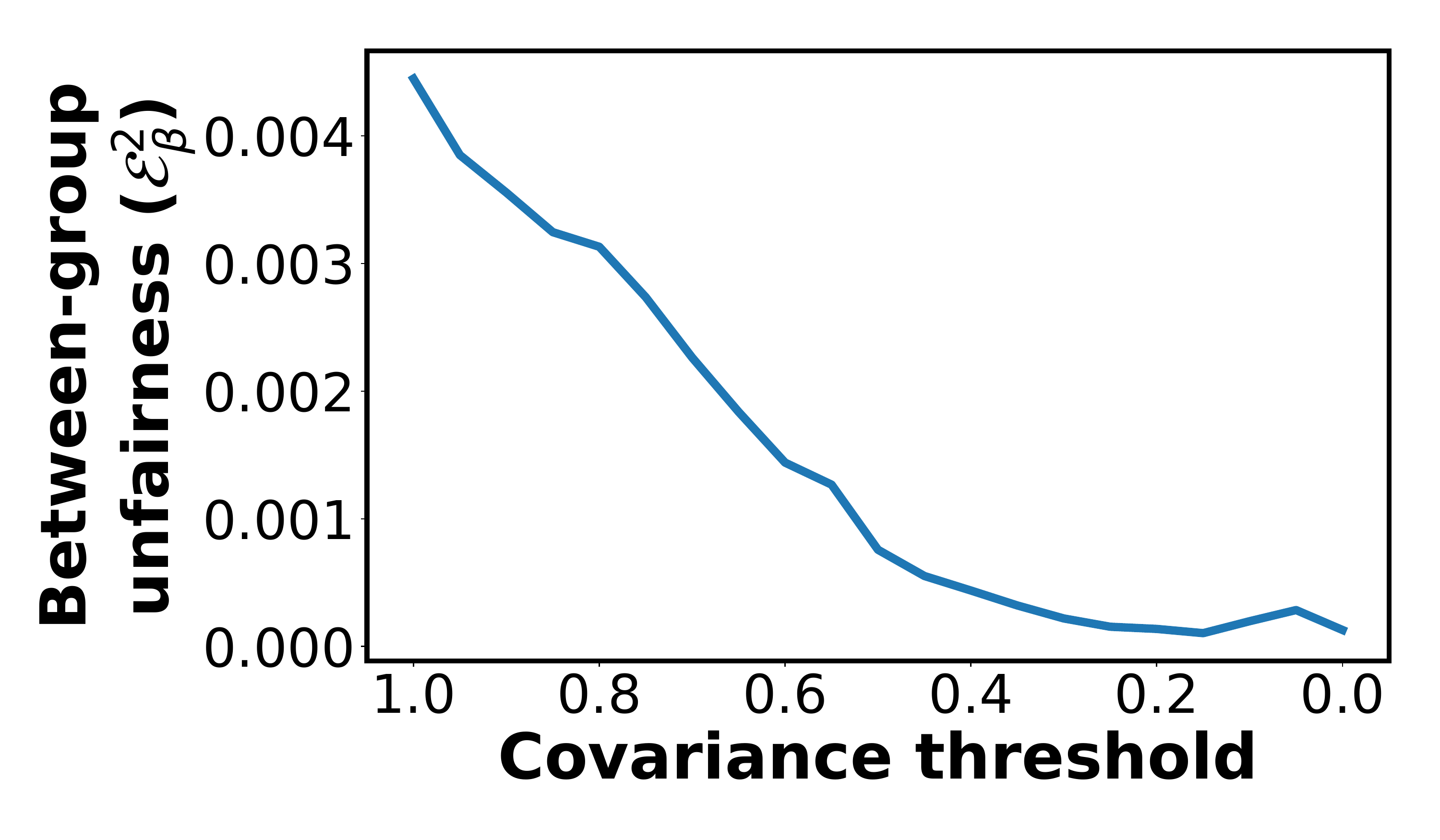}
		               \end{subfigure}
            \begin{subfigure}[b]{.30\textwidth}
	            \centering
                \hspace{11mm}
	            \textbf{Within-group}
                \includegraphics[width=\textwidth]{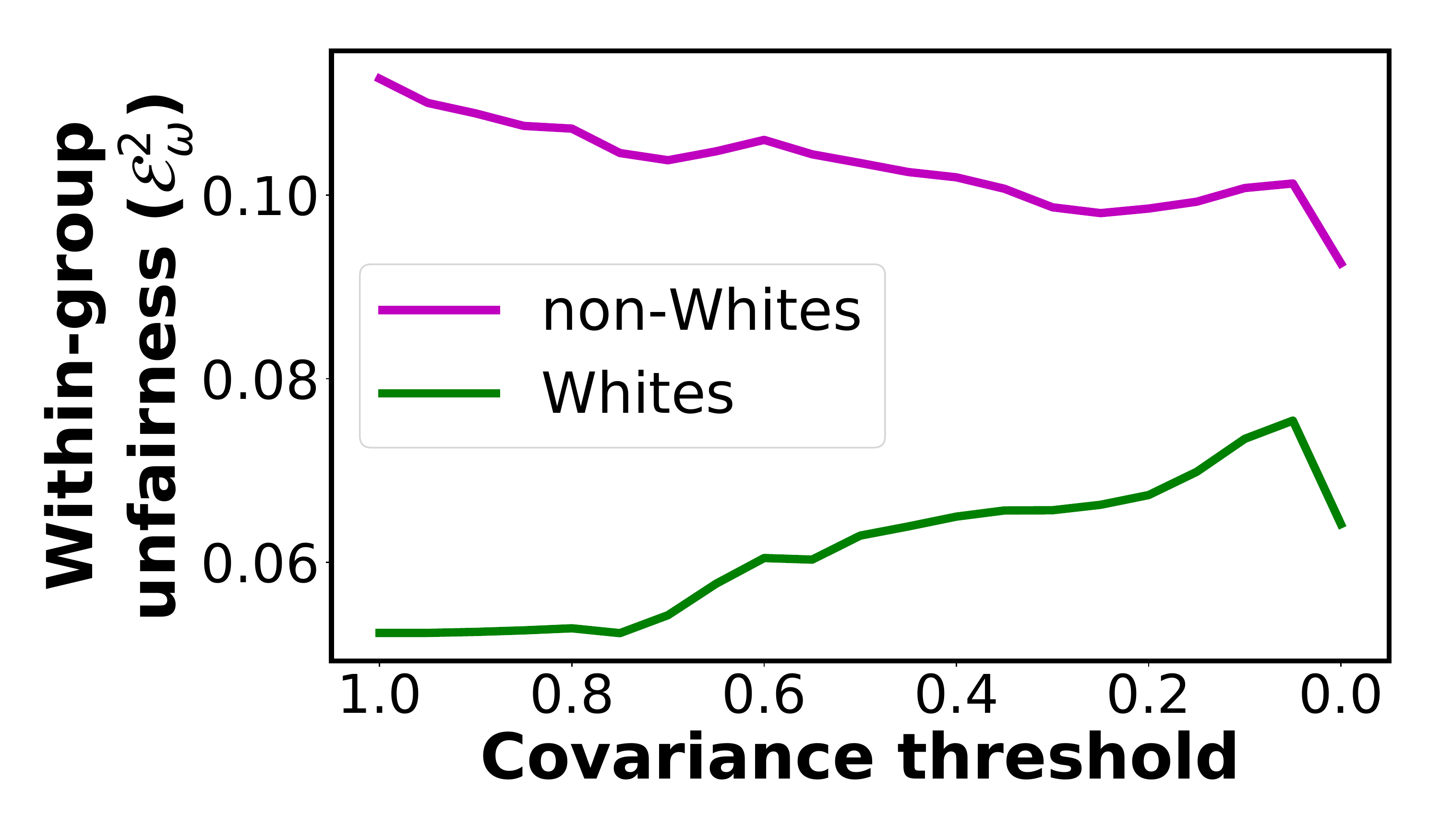}
		               \end{subfigure}
            \begin{subfigure}[b]{.30\textwidth}
	            \centering
                \hspace{9mm}
	            \textbf{Overall / individual}
                \includegraphics[width=\textwidth]{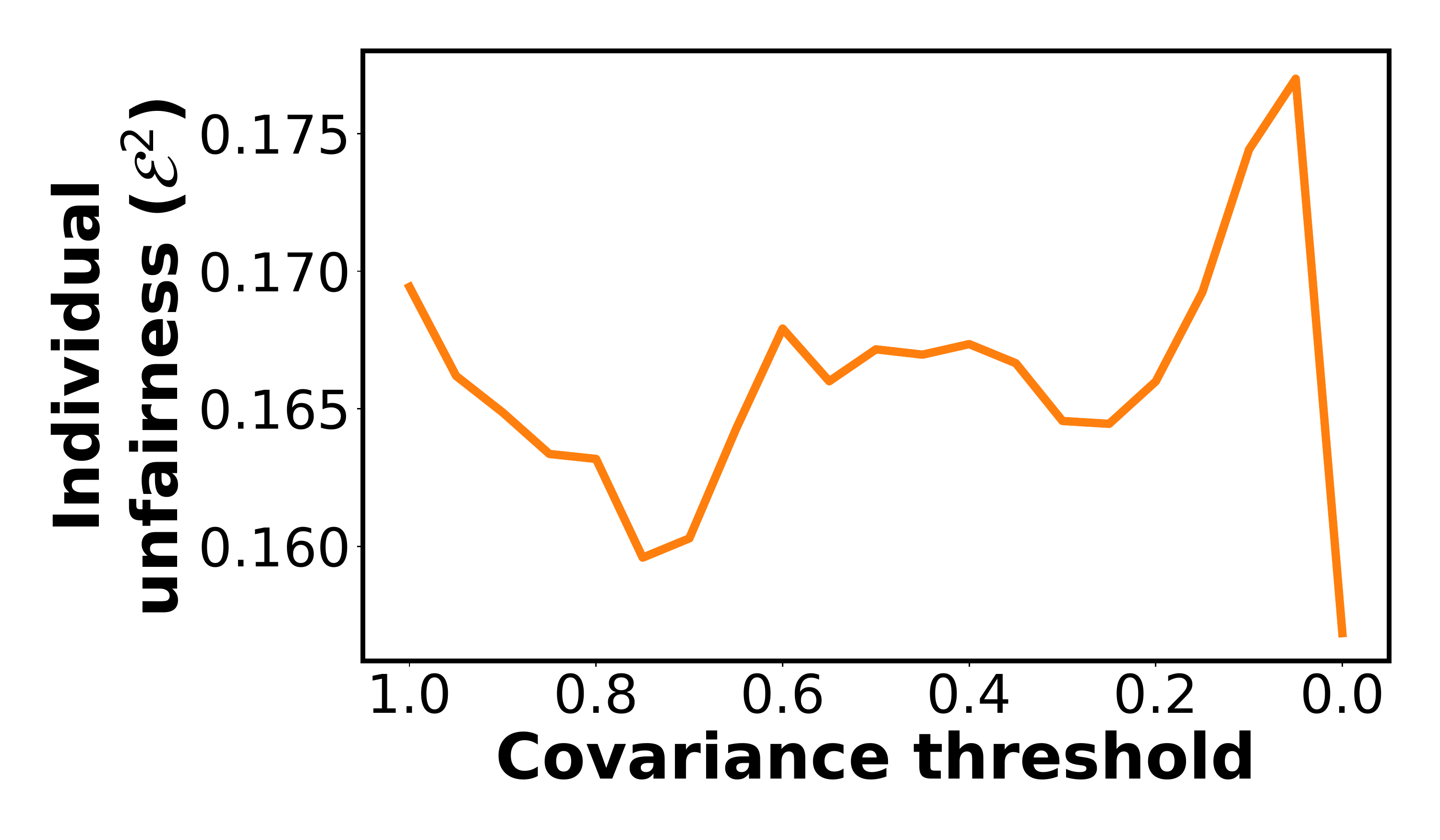}
		               \end{subfigure}            
                                                        \end{subfigure}
    \vspace{-7mm}
    \caption{\small{The effect of applying false negative rate constraints of Zafar et al. in order to minimize the between-group unfairness ($\Ecal^2_\beta(\veb)$). The second plot shows that reducing the between-group unfairness leads to an increase in the within-group unfairness ($\Ecal^2_\omega(\veb)$) for Whites. Moreover, for certain values of the covariance threshold, the overall unfairness ($\Ecal^2(\veb)$) increases as compared to the unconstrained classifier.}}        \label{fig:fair_const}
    \vspace{-3mm}
\end{figure*}

To test this hypothesis, we experiment with the following setup: 
We take the three sensitive features present in the COMPAS dataset, namely gender, race and age, and form sensitive feature groups based on all possible combinations of these features. For example, groups formed based on gender would be men and women; groups formed based on race would be Black, White and Hispanic; whereas groups formed based on gender as well as race would be Black men, Black women, White men and so on.
For each of these sensitive feature combinations we plot in Figure~\ref{fig:frac_btwn_grp} the percentage of contribution that the between-group unfairness has towards the overall unfairness. Figure~\ref{fig:frac_btwn_grp} shows that \emph{as the number of sensitive feature groups increases, the between-group unfairness contributes more and more towards the overall unfairness.} 
This result is also in line with the implications of Proposition~\ref{prop:intersection}.

Figure~\ref{fig:frac_btwn_grp} also shows the following interesting insight: Even though the overall unfairness and accuracy of all the classifiers is very similar  (cf. Figure~\ref{fig:indiv_accuracy}), the random forest classifier leads of significantly smaller contribution of between-group unfairness as compared to other classifier (\eg, gender, gender+age).
In other words even for similar levels of accuracy and overall unfairness, \emph{classifiers have very different between-group unfairness across different feature sets}.

\xhdr{Accuracy and between-group unfairness} 
We also study the between group unfairness in Figure~\ref{fig:within-group-accuracy} and corresponding accuracy in Figure~\ref{fig:indiv_accuracy}. 
We notice that there is \emph{no definitive correlation between the accuracy and the between-group unfairness}: In the Adult dataset, the highest level of accuracy (around $\tau=0.8$) corresponds to one of the lowest values of between-group unfairness for all classifiers. However, this doesn't hold in the case of COMPAS dataset.

\subsection{Interaction Between Different Types of Unfairness}\label{sec:empirical:interaction}

In this section, we revisit the literature on fairness-aware machine learning and investigate how methods proposed to control between-group unfairness (which is what most existing methods focus on \cite{zafar_fairness,zafar_dmt,feldman_kdd15,kamiran_classifying,kamishima_rec,icml2013_zemel13,hardt_nips16}) can affect the overall/individual and the within-group unfairness.
Specifically, we study how the overall unfairness ($\Ecal^2(\veb)$  in Eq.~\ref{eq:empirical:gr2}) and the within-group unfairness ($\Ecal^2_\omega(\veb)$  in Eq.~\ref{eq:empirical:gr2}) would change when training a constrained classifier to minimize the between-group unfairness ($\Ecal^2_\beta(\veb)$  in Eq.~\ref{eq:empirical:gr2}). 
Our study is motivated by the fact that while  several methods focus on designing constraints to remove the between-group unfairness (\eg, see~\cite{zafar_dmt,hardt_nips16,icml2013_zemel13}), to the best of our knowledge, no prior work in fairness-aware machine learning has studied the effect of these constraints on the overall and the within-group unfairness.

To this end, we use the methodology proposed by Zafar et al.~\cite{zafar_dmt} to remove the between-group unfairness based on false negative rates between different races (Whites and non-Whites) in the COMPAS dataset. Zafar et al. propose to remove the between-group unfairness by bounding the covariance between misclassification distance from the decision boundary and the sensitive feature value. The method operates by bounding the covariance of the unconstrained classifier by successive multiplicative factors between 1 and 0. A covariance multiplicative factor of $1$ means that no fairness constraints are applied while training the classifier, whereas a factor of $0$ means the tightest possible constraints are applied.
As done by Zafar et al., we train several logistic regression classifiers to limit the between-group unfairness; each classifier is trained with a covariance multiplicative factor in the range $[1.00, 0.95, 0.90, \ldots, 0.05, 0.00]$.

Figure~\ref{fig:fair_const} shows the between-group unfairness, within-group unfairness, and overall/individual unfairness as the fairness constraints of Zafar et al. \cite{zafar_dmt} are tightened towards $0$.
The figure shows the following key insights: (i) \emph{Reducing the between-group unfairness can in fact increase the within-group unfairness}: the within-group unfairness for Whites almost monotonically increases as the between-group unfairness is reduced. This observation also follows Proposition~\ref{prop:group}.
(ii) \emph{Reducing the between-group unfairness can exacerbate overall/individual unfairness}: 
 As the between-group unfairness decreases between the covariance multiplicative factor of $0.8$ to $0.6$ (on the x-axis), the overall unfairness in fact goes up.
 These insights point to possible significant tensions between these different components of unfairness.

\xhdr{Summary of empirical analysis}
Experiments on multiple real-world datasets performed in this section support the theoretical analysis of Section~\ref{sec:theoratical}. The empirical (as well as the theoretical) analysis brings out the inherent tensions between fairness and accuracy, as well as between different (between- and within-group) components of fairness. These results point to potential for situations where optimizing for one type of fairness can exacerbate the other.

\vspace{-0.5mm}
\section {Conclusion} \label{sec:discussion}
We proposed using inequality indices from economics as a principled way to compute the scalar degree of total unfairness of any algorithmic decision system. The approach is based on well-justified principles (axioms), and is general enough so that by varying the benefit function, we can capture all previous notions of algorithmic fairness conditions as special cases, while also admitting interesting generalizations. The resulting measures of total unfairness unify previous concepts of group and individual fairness, and allow us to study quantitatively the behavior of earlier methods to mitigate unfairness. These earlier methods typically worry only about between-group unfairness, which may be justified for legal reasons, or in order to redress particular social prejudices. However, we demonstrate that minimizing exclusively between-group unfairness may actually increase overall unfairness.

\section*{Acknowledgements} \label{sec:acknowledgements}

AW acknowledges support from the David MacKay Newton research fellowship at Darwin College, The Alan Turing Institute under EPSRC grant EP/N510129/1 \& TU/B/000074, and the Leverhulme Trust via the CFI.
HH acknowledges support from the Innosuisse grant 27248.1 PFES-ES.

\bibliographystyle{ACM-Reference-Format}
\balance
\bibliography{inequality_indices}


\begin{thebibliography}{42}


\ifx \showCODEN    \undefined \def \showCODEN     #1{\unskip}     \fi
\ifx \showDOI      \undefined \def \showDOI       #1{#1}\fi
\ifx \showISBNx    \undefined \def \showISBNx     #1{\unskip}     \fi
\ifx \showISBNxiii \undefined \def \showISBNxiii  #1{\unskip}     \fi
\ifx \showISSN     \undefined \def \showISSN      #1{\unskip}     \fi
\ifx \showLCCN     \undefined \def \showLCCN      #1{\unskip}     \fi
\ifx \shownote     \undefined \def \shownote      #1{#1}          \fi
\ifx \showarticletitle \undefined \def \showarticletitle #1{#1}   \fi
\ifx \showURL      \undefined \def \showURL       {\relax}        \fi
\providecommand\bibfield[2]{#2}
\providecommand\bibinfo[2]{#2}
\providecommand\natexlab[1]{#1}
\providecommand\showeprint[2][]{arXiv:#2}

\bibitem[\protect\citeauthoryear{Abdi}{Abdi}{2010}]%
        {abdi2010coefficient}
\bibfield{author}{\bibinfo{person}{Herv{\'e} Abdi}.}
  \bibinfo{year}{2010}\natexlab{}.
\newblock \showarticletitle{Coefficient of variation}.
\newblock \bibinfo{journal}{\emph{Encyclopedia of research design}}
  \bibinfo{volume}{1} (\bibinfo{year}{2010}), \bibinfo{pages}{169--171}.
\newblock


\bibitem[\protect\citeauthoryear{Angwin, Larson, Mattu, and Kirchner}{Angwin
  et~al\mbox{.}}{2016}]%
        {propublica_story}
\bibfield{author}{\bibinfo{person}{Julia Angwin}, \bibinfo{person}{Jeff
  Larson}, \bibinfo{person}{Surya Mattu}, {and} \bibinfo{person}{Lauren
  Kirchner}.} \bibinfo{year}{2016}\natexlab{}.
\newblock \bibinfo{title}{{Machine Bias: There's Software Used Across the
  Country to Predict Future Criminals. And it's Biased Against Blacks.}}
\newblock
  \bibinfo{howpublished}{\href{https://www.propublica.org/article/machine-bias-risk-assessments-in-criminal-sentencing}{https://www.propublica.org/article/machine-bias-risk-assessments-in-criminal-sentencing}}.
\newblock


\bibitem[\protect\citeauthoryear{Atkinson}{Atkinson}{1970}]%
        {atkinson1970measurement}
\bibfield{author}{\bibinfo{person}{Anthony~B. Atkinson}.}
  \bibinfo{year}{1970}\natexlab{}.
\newblock \showarticletitle{On the measurement of inequality}.
\newblock \bibinfo{journal}{\emph{Journal of economic theory}}
  \bibinfo{volume}{2}, \bibinfo{number}{3} (\bibinfo{year}{1970}),
  \bibinfo{pages}{244--263}.
\newblock


\bibitem[\protect\citeauthoryear{Barocas and Selbst}{Barocas and
  Selbst}{2016}]%
        {barocas2016big}
\bibfield{author}{\bibinfo{person}{Solon Barocas} {and}
  \bibinfo{person}{Andrew~D Selbst}.} \bibinfo{year}{2016}\natexlab{}.
\newblock \showarticletitle{Big data's disparate impact}.
\newblock \bibinfo{journal}{\emph{California Law Review}}
  \bibinfo{volume}{104} (\bibinfo{year}{2016}), \bibinfo{pages}{671}.
\newblock


\bibitem[\protect\citeauthoryear{Bell{\`u} and Liberati}{Bell{\`u} and
  Liberati}{2006}]%
        {bellu2006inequality}
\bibfield{author}{\bibinfo{person}{Lorenzo~Giovanni Bell{\`u}} {and}
  \bibinfo{person}{Paolo Liberati}.} \bibinfo{year}{2006}\natexlab{}.
\newblock \showarticletitle{Inequality analysis: The gini index}.
\newblock \bibinfo{journal}{\emph{Food and Agriculture Organization of the
  United Nations, EASYPol Module}}  \bibinfo{volume}{40}
  (\bibinfo{year}{2006}).
\newblock


\bibitem[\protect\citeauthoryear{Berk, Heidari, Jabbari, Kearns, and Roth}{Berk
  et~al\mbox{.}}{2017}]%
        {berk2017fairness}
\bibfield{author}{\bibinfo{person}{Richard Berk}, \bibinfo{person}{Hoda
  Heidari}, \bibinfo{person}{Shahin Jabbari}, \bibinfo{person}{Michael Kearns},
  {and} \bibinfo{person}{Aaron Roth}.} \bibinfo{year}{2017}\natexlab{}.
\newblock \showarticletitle{Fairness in criminal justice risk assessments: The
  state of the art}.
\newblock \bibinfo{journal}{\emph{arXiv preprint arXiv:1703.09207}}
  (\bibinfo{year}{2017}).
\newblock


\bibitem[\protect\citeauthoryear{Chouldechova}{Chouldechova}{2016}]%
        {dimpact_fpr}
\bibfield{author}{\bibinfo{person}{Alexandra Chouldechova}.}
  \bibinfo{year}{2016}\natexlab{}.
\newblock \showarticletitle{{Fair Prediction with Disparate Impact: A Study of
  Bias in Recidivism Prediction Instruments}}.
\newblock \bibinfo{journal}{\emph{{arXiv:1610.07524}}} (\bibinfo{year}{2016}).
\newblock


\bibitem[\protect\citeauthoryear{Concei{\c{c}}{\~a}o and
  Ferreira}{Concei{\c{c}}{\~a}o and Ferreira}{2000}]%
        {conceiccao2000young}
\bibfield{author}{\bibinfo{person}{Pedro Concei{\c{c}}{\~a}o} {and}
  \bibinfo{person}{Pedro Ferreira}.} \bibinfo{year}{2000}\natexlab{}.
\newblock \showarticletitle{The young person's guide to the Theil index:
  Suggesting intuitive interpretations and exploring analytical applications}.
\newblock  (\bibinfo{year}{2000}).
\newblock


\bibitem[\protect\citeauthoryear{Corbett{-}Davies, Pierson, Feller, Goel, and
  Huq}{Corbett{-}Davies et~al\mbox{.}}{2017}]%
        {Corbett-DaviesP17}
\bibfield{author}{\bibinfo{person}{Sam Corbett{-}Davies}, \bibinfo{person}{Emma
  Pierson}, \bibinfo{person}{Avi Feller}, \bibinfo{person}{Sharad Goel}, {and}
  \bibinfo{person}{Aziz Huq}.} \bibinfo{year}{2017}\natexlab{}.
\newblock \showarticletitle{Algorithmic decision making and the cost of
  fairness}. In \bibinfo{booktitle}{\emph{Proceedings of KDD}}.
\newblock


\bibitem[\protect\citeauthoryear{Cowell and Kuga}{Cowell and Kuga}{1981}]%
        {cowell1981additivity}
\bibfield{author}{\bibinfo{person}{Frank~A. Cowell} {and}
  \bibinfo{person}{Kiyoshi Kuga}.} \bibinfo{year}{1981}\natexlab{}.
\newblock \showarticletitle{Additivity and the entropy concept: an axiomatic
  approach to inequality measurement}.
\newblock \bibinfo{journal}{\emph{Journal of Economic Theory}}
  \bibinfo{volume}{25}, \bibinfo{number}{1} (\bibinfo{year}{1981}),
  \bibinfo{pages}{131--143}.
\newblock


\bibitem[\protect\citeauthoryear{Dalton}{Dalton}{1920}]%
        {dalton1920measurement}
\bibfield{author}{\bibinfo{person}{Hugh Dalton}.}
  \bibinfo{year}{1920}\natexlab{}.
\newblock \showarticletitle{The measurement of the inequality of incomes}.
\newblock \bibinfo{journal}{\emph{The Economic Journal}} \bibinfo{volume}{30},
  \bibinfo{number}{119} (\bibinfo{year}{1920}), \bibinfo{pages}{348--361}.
\newblock


\bibitem[\protect\citeauthoryear{Dwork, Hardt, Pitassi, Reingold, and
  Zemel}{Dwork et~al\mbox{.}}{2012}]%
        {dwork2012fairness}
\bibfield{author}{\bibinfo{person}{Cynthia Dwork}, \bibinfo{person}{Moritz
  Hardt}, \bibinfo{person}{Toniann Pitassi}, \bibinfo{person}{Omer Reingold},
  {and} \bibinfo{person}{Richard Zemel}.} \bibinfo{year}{2012}\natexlab{}.
\newblock \showarticletitle{Fairness through awareness}. In
  \bibinfo{booktitle}{\emph{Proceedings of the 3rd Innovations in Theoretical
  Computer Science Conference (ITCS)}}. \bibinfo{pages}{214--226}.
\newblock


\bibitem[\protect\citeauthoryear{Feldman, Friedler, Moeller, Scheidegger, and
  Venkatasubramanian}{Feldman et~al\mbox{.}}{2015}]%
        {feldman_kdd15}
\bibfield{author}{\bibinfo{person}{Michael Feldman},
  \bibinfo{person}{Sorelle~A. Friedler}, \bibinfo{person}{John Moeller},
  \bibinfo{person}{Carlos Scheidegger}, {and} \bibinfo{person}{Suresh
  Venkatasubramanian}.} \bibinfo{year}{2015}\natexlab{}.
\newblock \showarticletitle{Certifying and Removing Disparate Impact}. In
  \bibinfo{booktitle}{\emph{Proceedings of KDD}}.
\newblock


\bibitem[\protect\citeauthoryear{Flores, Lowenkamp, and Bechtel}{Flores
  et~al\mbox{.}}{2016}]%
        {flores_propublica_reply}
\bibfield{author}{\bibinfo{person}{Anthony~W. Flores},
  \bibinfo{person}{Christopher~T. Lowenkamp}, {and} \bibinfo{person}{Kristin
  Bechtel}.} \bibinfo{year}{2016}\natexlab{}.
\newblock \showarticletitle{{False Positives, False Negatives, and False
  Analyses: A Rejoinder to "Machine Bias: There's Software Used Across the
  Country to Predict Future Criminals. And it's Biased Against Blacks."}}.
\newblock  (\bibinfo{year}{2016}).
\newblock


\bibitem[\protect\citeauthoryear{Furletti}{Furletti}{2002}]%
        {history-of-credit}
\bibfield{author}{\bibinfo{person}{Mark~J. Furletti}.}
  \bibinfo{year}{2002}\natexlab{}.
\newblock \bibinfo{title}{An Overview and History of Credit Reporting}.
\newblock
\newblock
\newblock
\shownote{\url{http://dx.doi.org/10.2139/ssrn.927487}.}


\bibitem[\protect\citeauthoryear{Gini}{Gini}{1912}]%
        {gini1912variabilita}
\bibfield{author}{\bibinfo{person}{Corrado Gini}.}
  \bibinfo{year}{1912}\natexlab{}.
\newblock \showarticletitle{Variabilit{\`a} e mutabilit{\`a}}.
\newblock \bibinfo{journal}{\emph{Reprinted in Memorie di metodologica
  statistica (Ed. Pizetti E, Salvemini, T). Rome: Libreria Eredi Virgilio
  Veschi}} (\bibinfo{year}{1912}).
\newblock


\bibitem[\protect\citeauthoryear{Hardt, Price, Srebro, et~al\mbox{.}}{Hardt
  et~al\mbox{.}}{2016}]%
        {hardt_nips16}
\bibfield{author}{\bibinfo{person}{Moritz Hardt}, \bibinfo{person}{Eric Price},
  \bibinfo{person}{Nati Srebro}, {et~al\mbox{.}}}
  \bibinfo{year}{2016}\natexlab{}.
\newblock \showarticletitle{Equality of opportunity in supervised learning}. In
  \bibinfo{booktitle}{\emph{Proceedings of NIPS}}. \bibinfo{pages}{3315--3323}.
\newblock


\bibitem[\protect\citeauthoryear{Joseph, Kearns, Morgenstern, and Roth}{Joseph
  et~al\mbox{.}}{2016}]%
        {joseph2016fairness}
\bibfield{author}{\bibinfo{person}{Matthew Joseph}, \bibinfo{person}{Michael
  Kearns}, \bibinfo{person}{Jamie Morgenstern}, {and} \bibinfo{person}{Aaron
  Roth}.} \bibinfo{year}{2016}\natexlab{}.
\newblock \showarticletitle{Fairness in learning: Classic and contextual
  bandits}. In \bibinfo{booktitle}{\emph{Proceedings of NIPS}}.
  \bibinfo{pages}{325--333}.
\newblock


\bibitem[\protect\citeauthoryear{Kakwani}{Kakwani}{1980}]%
        {kakwani1980class}
\bibfield{author}{\bibinfo{person}{Nanak Kakwani}.}
  \bibinfo{year}{1980}\natexlab{}.
\newblock \showarticletitle{On a class of poverty measures}.
\newblock \bibinfo{journal}{\emph{Econometrica: Journal of the Econometric
  Society}} (\bibinfo{year}{1980}), \bibinfo{pages}{437--446}.
\newblock


\bibitem[\protect\citeauthoryear{Kamiran and Calders}{Kamiran and
  Calders}{2009}]%
        {kamiran_classifying}
\bibfield{author}{\bibinfo{person}{Faisal Kamiran} {and} \bibinfo{person}{Toon
  Calders}.} \bibinfo{year}{2009}\natexlab{}.
\newblock \showarticletitle{Classifying without discriminating}. In
  \bibinfo{booktitle}{\emph{Proceedings of the 2nd International Conference on
  Computer, Control and Communication}}. IEEE, \bibinfo{pages}{1--6}.
\newblock


\bibitem[\protect\citeauthoryear{Kamishima, Akaho, and Sakuma}{Kamishima
  et~al\mbox{.}}{2013}]%
        {kamishima_rec}
\bibfield{author}{\bibinfo{person}{Toshihiro Kamishima},
  \bibinfo{person}{Shotaro Akaho}, {and} \bibinfo{person}{Jun Sakuma}.}
  \bibinfo{year}{2013}\natexlab{}.
\newblock \showarticletitle{Efficiency improvement of neutrality-enhanced
  recommendation}. In \bibinfo{booktitle}{\emph{{RecSys}}}.
\newblock


\bibitem[\protect\citeauthoryear{Kearns, Neel, Roth, and Wu}{Kearns
  et~al\mbox{.}}{2017}]%
        {kearns2017preventing}
\bibfield{author}{\bibinfo{person}{Michael Kearns}, \bibinfo{person}{Seth
  Neel}, \bibinfo{person}{Aaron Roth}, {and} \bibinfo{person}{Zhiwei~Steven
  Wu}.} \bibinfo{year}{2017}\natexlab{}.
\newblock \showarticletitle{Preventing fairness gerrymandering: Auditing and
  learning for subgroup fairness}.
\newblock \bibinfo{journal}{\emph{arXiv preprint arXiv:1711.05144}}
  (\bibinfo{year}{2017}).
\newblock


\bibitem[\protect\citeauthoryear{Kleinberg, Mullainathan, and
  Raghavan}{Kleinberg et~al\mbox{.}}{2017}]%
        {klein16}
\bibfield{author}{\bibinfo{person}{Jon Kleinberg}, \bibinfo{person}{Sendhil
  Mullainathan}, {and} \bibinfo{person}{Manish Raghavan}.}
  \bibinfo{year}{2017}\natexlab{}.
\newblock \showarticletitle{{Inherent Trade-Offs in the Fair Determination of
  Risk Scores}}. In \bibinfo{booktitle}{\emph{Proceedings of ITCS}}.
\newblock


\bibitem[\protect\citeauthoryear{Kolm}{Kolm}{1976a}]%
        {kolm1976unequal}
\bibfield{author}{\bibinfo{person}{Serge-Christophe Kolm}.}
  \bibinfo{year}{1976}\natexlab{a}.
\newblock \showarticletitle{Unequal inequalities. I}.
\newblock \bibinfo{journal}{\emph{Journal of economic Theory}}
  \bibinfo{volume}{12}, \bibinfo{number}{3} (\bibinfo{year}{1976}),
  \bibinfo{pages}{416--442}.
\newblock


\bibitem[\protect\citeauthoryear{Kolm}{Kolm}{1976b}]%
        {kolm1976unequali}
\bibfield{author}{\bibinfo{person}{Serge-Christophe Kolm}.}
  \bibinfo{year}{1976}\natexlab{b}.
\newblock \showarticletitle{Unequal inequalities. II}.
\newblock \bibinfo{journal}{\emph{Journal of Economic Theory}}
  \bibinfo{volume}{13}, \bibinfo{number}{1} (\bibinfo{year}{1976}),
  \bibinfo{pages}{82--111}.
\newblock


\bibitem[\protect\citeauthoryear{Larson, Mattu, Kirchner, and Angwin}{Larson
  et~al\mbox{.}}{2016}]%
        {propublica_data}
\bibfield{author}{\bibinfo{person}{Jeff Larson}, \bibinfo{person}{Surya Mattu},
  \bibinfo{person}{Lauren Kirchner}, {and} \bibinfo{person}{Julia Angwin}.}
  \bibinfo{year}{2016}\natexlab{}.
\newblock \bibinfo{title}{Data and analysis for `{How} we analyzed the {COMPAS}
  recidivism algorithm'}.
\newblock
  \bibinfo{howpublished}{\url{https://github.com/propublica/compas-analysis}}.
\newblock


\bibitem[\protect\citeauthoryear{Lichman}{Lichman}{2013}]%
        {adult_dataset}
\bibfield{author}{\bibinfo{person}{M. Lichman}.}
  \bibinfo{year}{2013}\natexlab{}.
\newblock \bibinfo{title}{{UCI} machine learning repository: The Adult income
  data set}.
\newblock
  \bibinfo{howpublished}{\url{https://archive.ics.uci.edu/ml/datasets/adult}}.
\newblock


\bibitem[\protect\citeauthoryear{Litchfield}{Litchfield}{1999}]%
        {litchfield1999inequality}
\bibfield{author}{\bibinfo{person}{Julie~A. Litchfield}.}
  \bibinfo{year}{1999}\natexlab{}.
\newblock \showarticletitle{Inequality: Methods and tools}.
\newblock \bibinfo{journal}{\emph{World Bank}}  \bibinfo{volume}{4}
  (\bibinfo{year}{1999}).
\newblock


\bibitem[\protect\citeauthoryear{Long and Nucci}{Long and Nucci}{1997}]%
        {long1997hoover}
\bibfield{author}{\bibinfo{person}{Larry Long} {and} \bibinfo{person}{Alfred
  Nucci}.} \bibinfo{year}{1997}\natexlab{}.
\newblock \showarticletitle{The Hoover index of population concentration: A
  correction and update}.
\newblock \bibinfo{journal}{\emph{The Professional Geographer}}
  (\bibinfo{year}{1997}).
\newblock


\bibitem[\protect\citeauthoryear{Lundberg and Lee}{Lundberg and Lee}{2017}]%
        {lundberg2017unified}
\bibfield{author}{\bibinfo{person}{Scott~M. Lundberg} {and}
  \bibinfo{person}{Su-In Lee}.} \bibinfo{year}{2017}\natexlab{}.
\newblock \showarticletitle{A unified approach to interpreting model
  predictions}. In \bibinfo{booktitle}{\emph{Proceedings of NIPS}}.
  \bibinfo{pages}{4768--4777}.
\newblock


\bibitem[\protect\citeauthoryear{Pigou}{Pigou}{1912}]%
        {pigou1912wealth}
\bibfield{author}{\bibinfo{person}{Arthur~Cecil Pigou}.}
  \bibinfo{year}{1912}\natexlab{}.
\newblock \bibinfo{booktitle}{\emph{Wealth and welfare}}.
\newblock \bibinfo{publisher}{Macmillan and Company, limited}.
\newblock


\bibitem[\protect\citeauthoryear{Podesta, Pritzker, Moniz, Holdren, and
  Zients}{Podesta et~al\mbox{.}}{2014}]%
        {bigdatawhitehouse}
\bibfield{author}{\bibinfo{person}{John Podesta}, \bibinfo{person}{Penny
  Pritzker}, \bibinfo{person}{Ernest Moniz}, \bibinfo{person}{John Holdren},
  {and} \bibinfo{person}{Jeffrey Zients}.} \bibinfo{year}{2014}\natexlab{}.
\newblock \showarticletitle{Big data: Seizing opportunities, preserving
  values}.
\newblock \bibinfo{journal}{\emph{Executive Office of the President. The White
  House.}} (\bibinfo{year}{2014}).
\newblock


\bibitem[\protect\citeauthoryear{Romei and Ruggieri}{Romei and
  Ruggieri}{2014}]%
        {salvatore_survey}
\bibfield{author}{\bibinfo{person}{Andrea Romei} {and}
  \bibinfo{person}{Salvatore Ruggieri}.} \bibinfo{year}{2014}\natexlab{}.
\newblock \showarticletitle{A multidisciplinary survey on discrimination
  analysis}.
\newblock \bibinfo{journal}{\emph{The Knowledge Engineering Review}}
  \bibinfo{volume}{29}, \bibinfo{number}{5} (\bibinfo{year}{2014}),
  \bibinfo{pages}{582--638}.
\newblock


\bibitem[\protect\citeauthoryear{Sen}{Sen}{1973}]%
        {sen1973economic}
\bibfield{author}{\bibinfo{person}{Amartya Sen}.}
  \bibinfo{year}{1973}\natexlab{}.
\newblock \bibinfo{booktitle}{\emph{On economic inequality}}.
\newblock \bibinfo{publisher}{Oxford University Press}.
\newblock


\bibitem[\protect\citeauthoryear{Shorrocks}{Shorrocks}{1980}]%
        {shorrocks1980class}
\bibfield{author}{\bibinfo{person}{Anthony~F Shorrocks}.}
  \bibinfo{year}{1980}\natexlab{}.
\newblock \showarticletitle{The class of additively decomposable inequality
  measures}.
\newblock \bibinfo{journal}{\emph{Econometrica: Journal of the Econometric
  Society}} (\bibinfo{year}{1980}), \bibinfo{pages}{613--625}.
\newblock


\bibitem[\protect\citeauthoryear{Subramanian}{Subramanian}{2011}]%
        {subramanian2011inequality}
\bibfield{author}{\bibinfo{person}{Subbu Subramanian}.}
  \bibinfo{year}{2011}\natexlab{}.
\newblock \showarticletitle{Inequality measurement with subgroup
  decomposability and level-sensitivity}.
\newblock  (\bibinfo{year}{2011}).
\newblock


\bibitem[\protect\citeauthoryear{Summers and Willis}{Summers and
  Willis}{2010}]%
        {summers2010pretrial}
\bibfield{author}{\bibinfo{person}{Charles Summers} {and} \bibinfo{person}{Tim
  Willis}.} \bibinfo{year}{2010}\natexlab{}.
\newblock \bibinfo{title}{Pretrial risk assessment}.
\newblock
\newblock


\bibitem[\protect\citeauthoryear{Sundararajan, Taly, and Yan}{Sundararajan
  et~al\mbox{.}}{2017}]%
        {Sun17}
\bibfield{author}{\bibinfo{person}{Mukund Sundararajan}, \bibinfo{person}{Ankur
  Taly}, {and} \bibinfo{person}{Qiqi Yan}.} \bibinfo{year}{2017}\natexlab{}.
\newblock \showarticletitle{Axiomatic Attribution for Deep Networks}. In
  \bibinfo{booktitle}{\emph{Proceedings of ICML}}. \bibinfo{pages}{3319--3328}.
\newblock


\bibitem[\protect\citeauthoryear{Theil}{Theil}{1967}]%
        {theil1967}
\bibfield{author}{\bibinfo{person}{Henri Theil}.}
  \bibinfo{year}{1967}\natexlab{}.
\newblock \bibinfo{booktitle}{\emph{Economics and Information Theory}}.
\newblock \bibinfo{publisher}{North Holland}, \bibinfo{address}{Amsterdam.}
\newblock


\bibitem[\protect\citeauthoryear{Zafar, Valera, Rodriguez, and Gummadi}{Zafar
  et~al\mbox{.}}{2017a}]%
        {zafar_dmt}
\bibfield{author}{\bibinfo{person}{Muhammad~Bilal Zafar},
  \bibinfo{person}{Isabel Valera}, \bibinfo{person}{Manuel~Gomez Rodriguez},
  {and} \bibinfo{person}{Krishna~P. Gummadi}.}
  \bibinfo{year}{2017}\natexlab{a}.
\newblock \showarticletitle{Fairness beyond disparate treatment \& disparate
  impact: Learning classification without disparate mistreatment}. In
  \bibinfo{booktitle}{\emph{In proceedings of WWW}}.
\newblock


\bibitem[\protect\citeauthoryear{Zafar, Valera, Rodriguez, and Gummadi}{Zafar
  et~al\mbox{.}}{2017b}]%
        {zafar_fairness}
\bibfield{author}{\bibinfo{person}{Muhammad~Bilal Zafar},
  \bibinfo{person}{Isabel Valera}, \bibinfo{person}{Manuel~Gomez Rodriguez},
  {and} \bibinfo{person}{Krishna~P. Gummadi}.}
  \bibinfo{year}{2017}\natexlab{b}.
\newblock \showarticletitle{Fairness constraints: Mechanisms for fair
  classification}. In \bibinfo{booktitle}{\emph{Proceedings of AISTATS}}.
\newblock


\bibitem[\protect\citeauthoryear{Zemel, Wu, Swersky, Pitassi, and Dwork}{Zemel
  et~al\mbox{.}}{2013}]%
        {icml2013_zemel13}
\bibfield{author}{\bibinfo{person}{Rich Zemel}, \bibinfo{person}{Yu Wu},
  \bibinfo{person}{Kevin Swersky}, \bibinfo{person}{Toni Pitassi}, {and}
  \bibinfo{person}{Cynthia Dwork}.} \bibinfo{year}{2013}\natexlab{}.
\newblock \showarticletitle{Learning fair representations}. In
  \bibinfo{booktitle}{\emph{Proceedings of ICML}}. \bibinfo{pages}{325--333}.
\newblock


\end{thebibliography}

\shortLongVersion{}{
    \pagebreak
    \appendix
    \section{Appendix: Technical Material}\label{app:technical}

\paragraph{Proof of Proposition~\ref{prop:bayes}}
If there exists a classifier $\theta'$ with $L_\Dcal(\theta') =0$, then that classifier minimizes $I$: For all $i=1,\cdots,n$, the benefit $i$ receives under $\theta'$, denoted by $b^{\theta'}_i$, is equal to $1 + \theta'(\vx_i) - y_i = 1$. That is everyone gets the same benefit under $\theta'$, and as the result, $I(\veb^{\theta'}) = I(\ones)=0$.

If there exists a classifier $\theta$ with $I(\veb^\theta)=0$, $\theta$ must assign the same benefit to everyone: there exists $b\in\{0,1,2\}$ such that for all $i=1,\cdots,n$, $b^\theta_i = 1 + \theta(\vx_i) - y_i = b$. Now let $\theta'(\vx) = \theta(\vx) + 1 - b$. It is easy to verify that $\theta'$ has zero error (i.e. $L_\Dcal(\theta') = 0$).
\qed

\paragraph{Proof of Proposition~\ref{prop:bayes2}}
Let $p = \Prob_{(\vx,y)\sim \cP}[y =1 \vert \vx=\tilde{\vx} ]$ so that $0<p<0.5$. 
We know that the Bayes/accuracy optimal classifier assigns label 0 to every instance $i$ with $\vx_i=\tilde{\vx}$. 
Suppose $\Dcal$ consists of $n$ instances all with $\vx_i=\tilde{\vx}$. Given the population invariance property of $I$, $n$ can be arbitrarily large without affecting the value of $I(.)$. Therefore we instead reason about the limiting case where $n=\infty$. In this case, we expect exactly $p$ fraction of the instances to have $y=1$ and the other $(1-p)$ fraction to have $y=0$. The benefit distribution $\veb^A$ for $\theta^A$, therefore, consists of $p$ fraction of the population receiving benefit $0$ and the other $(1-p)$ fraction receiving $1$.

Now consider a probabilistic classifier $\theta^q$ that randomly assigns label 1 to each instance in $\Dcal$ with probability $q \in (0,1)$. The resulting benefit distribution $\veb^q$ of $\theta^q$ is as follows: $p(1-q)$ fraction of the population receive benefit $0$; $pq+(1-p)(1-q)$ faction receive benefit $1$; and $(1-p)q$ faction receive benefit $2$. 

We claim that for $q=(1-p)$, $I_\Dcal(\theta^A) > I_\Dcal(\theta^q)$. To see this, note that $\veb^q$ can be constructed from $\veb^A$ via a series of inequality-reducing operations:
\begin{enumerate}
\item $\veb' = 2 \times \veb^A$, so that $\veb'$ consists of $p$ fraction of the population receiving benefit $0$ and the other $(1-p)$ fraction receiving $2$. Due to the scale invariance property of $I$, we know $I(\veb') = I(\veb^A)$.
\item Perform the following progressive transfer on $\veb'$ to obtain $\veb''$: Take one unit of benefit from $p(1-p)$ fraction of the population whose benefit is  2, and give it to $p(1-p)$ fraction with benefit 0. The resulting distribution, $\veb''$, consists of $p^2$ faction with benefit $0$; $2p(1-p)$ faction with benefit $1$; and $(1-p)^2$ faction with benefit $2$. Because $I$ satisfies the Dalton principle and $p(1-p)>0$, we have that $I(\veb'')<I(\veb')$.
\end{enumerate}
Combining the above two, we have $I(\veb'')<I(\veb^A)$.
It only remains to note that $\veb''$ is precisely $\veb^q$ for $q=(1-p)$. Therefore, we conclude $I(\veb'')= I(\veb^{(1-p)})<I(\veb^A)$.
This finishes the proof.
\qed

The following example shows that the accuracy of the fairness optimal classifier can be arbitrarily bad compared to that of the accuracy optimal classifier.
\begin{example}\label{ex:fair_accuracy}
Consider the example in the proof of Proposition~\ref{prop:bayes2}. Let $I$ be the generalized entropy with $\alpha=2$. We claim that the fairness optimal classifier is one that assigns label 1 to every instance. To see this, recall that under $\theta^q$, $p(1-q)$ fraction of the population receive benefit $0$; $pq+(1-p)(1-q)$ faction receive benefit $1$; and $(1-p)q$ faction receive benefit $2$. So the mean benefit $\mu$ is equal to $1-p+q$. Taking derivative with respect to $q$, we have
{\tiny
$$\frac{d}{d q} \left( (pq+(1-p)(1-q))\left(\frac{1}{1-p+q}\right)^2 + (1-p)q\left(\frac{2}{1-p+q}\right)^2\right)=0$$
}
$$\Rightarrow q^* = \frac{1-3p + 2p^2}{3-2p}$$
The derivative is positive for $q<q^*$ and negative for $q>q^*$. The minimum therefore happens at either $q=0$ or $q=1$. Given that for $0<p<1$, $\frac{1}{1-p} > \frac{4-3p}{(2-p)^2}$, we obtain that $q=1$ minimizes $I$. 

The fairness optimal classifier assigns label 1 to every instance resulting in accuracy $p$, whereas the accuracy optimal classifier can achieve accuracy $(1-p)$. The ratio $\frac{1-p}{p}$ can be arbitrarily large if $p$ is taken to be sufficiently small.
\end{example}

\paragraph{Proof of Proposition~\ref{prop:group}}
Note that because of the optimality of $\theta^*_\beta$ for (\ref{eq:gr}), if $I_\beta(\theta^*_\beta)\neq I_\beta(\theta^*)$, it must be the case that $I_\beta(\theta^*_\beta) < I_\beta(\theta^*)$. If $I(\theta^*_\beta)\leq I(\theta^*)$, then $\theta^*_\beta$ is an optimal solution to (\ref{eq:ind}), and $I_\beta(\theta^*_\beta) < I_\beta(\theta^*)$. This is a contradiction with the choice of $\theta^*$. If $I_\omega(\theta^*_\beta)\leq I_\omega(\theta^*)$, then combined with the fact that $I_\beta(\theta^*_\beta) < I_\beta(\theta^*)$, we have that $I(\theta^*_\beta) < I(\theta^*)$, which is a contradiction with the optimality of $\theta^*$ for (\ref{eq:ind}).
\qed

\newcommand{\vmu}{\boldsymbol{\mu}}

\paragraph{Proof of Proposition~\ref{prop:intersection}}
Suppose $|G|=m$ and $|G'|=m'.$
Let $\veb = \left(\veb^{(g_1,g'_1)}, \cdots, \veb^{(g_m,g'_{m'})}\right)$ where $\veb^{(g_i,g'_j)}$ specifies the benefit distribution for individuals in group $(g_i,g'_j)$ and $\vmu^{(g_i,g'_j)}$ specifies the distribution in which each individual in $(g_i,g'_j)$ receives the group's mean benefit.
Note that $I^{G \times G'}_\beta(\veb)$ can be written as 
\begin{eqnarray*}
&&I\left(\vmu^{(g_1,g'_1)}, \cdots, \vmu^{(g_m,g'_{m'})}\right) \\
&=& I^G_\beta\left(\vmu^{(g_1,g'_1)}, \cdots, \vmu^{(g_m,g'_{m'})}\right) +  I^G_\omega\left(\vmu^{(g_1,g'_1)}, \cdots, \vmu^{(g_m,g'_{m'})}\right) \\
&= & I\left(\vmu^{g_1}, \cdots, \vmu^{g_m}\right) +  I^G_\omega\left(\vmu^{(g_1,g'_1)}, \cdots, \vmu^{(g_m,g'_{m'})}\right) \\
&\geq & I\left(\vmu^{g_1}, \cdots, \vmu^{g_m}\right) \\
&= & I^G_\beta(\veb)
\end{eqnarray*}
where to obtain the second line, we used the additive decomposability property of $I$; for the third line we used the definition of the between-group component, and finally to obtain the conclusion, we used the zero-normalization property of $I$.
\qed

\paragraph{Proof of Proposition~\ref{prop:group_share}}
Recall that $I(\veb) = I^G_\beta(\veb) + I^G_\omega(\veb)$ and $I^G_\beta(\veb), I^G_\omega(\veb) \geq 0$. Therefore, we have $0 \leq \frac{I^G_\beta(\veb)}{I(\veb)} \leq 1$. 

Consider a benefit distribution $\veb$ in which members of group $g_1\in G$ receive benefit 1, and everyone else receives benefit $0$. It is easy to see that for this distribution $\frac{I^G_\beta(\veb)}{I(\veb)} = 1$. Similarly, consider a benefit distribution $\veb'$ that assigns a benefit of 1 to half of the population in each group, and 0 to everyone else. It is easy to that $\frac{I^G_\beta(\veb')}{I(\veb')} = 0$.
\qed

The following example shows that an added feature may in fact worsen the unfairness of the accuracy optimal classifier. 
\begin{example}\label{ex:feature}
Let $I(.)$ be the generalized entropy with $\alpha = 2$. Suppose for all $\vx_i=\tilde{\vx}$, $p = \Prob_{(\vx,y)\sim D}[y =1 \vert \vx=\tilde{\vx} ] >0.5$, so $\theta^A$ assigns label 1 to every instance with $\vx_i=\tilde{\vx}$. So in the resulting benefit distribution, $p$ fraction of the population receives benefit $1$ and the other $(1-p)$ fraction receives $2$. The mean benefit is, therefore, $(2-p)$ and GE is equal to\footnote{We are dropping the constants from the definition of the inequality index, as they don't affect the comparison.}
$$ p \left( \frac{1}{2-p} \right)^2 + (1-p)\left( \frac{2}{2-p} \right)^2 = \frac{4-3p}{(2-p)^2}.$$

Suppose with the addition of a new binary feature, the population breaks down into two subpopulations, one corresponding to $\vx=(\tilde{\vx},0)$ and the other corresponding to $\vx=(\tilde{\vx},1)$. Let $\frac{r}{1-r}$ be the relative size of the former sub-population to the latter ($0\leq r \leq 1$). We would like the accuracy optimal classifier to be different for each subpopulation, so for now let's assume:
\begin{itemize}
\item $\Prob_{(\vx,y)\sim D}[y =1 \wedge \vx=(\tilde{\vx},0) ] =\frac{r}{2} - \epsilon$, so $\theta^A$ assigns label 0 to every instance with $\vx=(\tilde{\vx},0)$---this is because $\frac{r}{2} - \epsilon < \frac{1}{2} r$.
\item $\Prob_{(\vx,y)\sim D}[y =1 \wedge \vx=(\tilde{\vx},1) ] =p-\frac{r}{2} + \epsilon$, so $\theta^A$ assigns label 1 to every instance with $\vx=(\tilde{\vx},1)$---this is because $p-\frac{r}{2} + \epsilon > \frac{1}{2} (1-r)$.
\end{itemize} 
In the resulting benefit distribution, $\frac{r}{2} -\epsilon$ fraction of the total population receives benefit $0$, $p+ 2\epsilon$ fraction of the total population receives benefit $1$, and the other $(1-p -\frac{r}{2} -\epsilon)$ fraction receives benefit $2$. The mean benefit is, therefore, $(2-p-r)$ and GE is equal to
{\small
\begin{eqnarray*}
&&(p+2\epsilon) \left( \frac{1}{2-p-r} \right)^2 + (1-p-\frac{r}{2} -\epsilon) \left( \frac{2}{2-p-r} \right)^2 \\
&&= \frac{4-3p-2r-2\epsilon}{(2-p-r)^2}.
\end{eqnarray*}
}
Take $p=0.9$,  $r=0.2$ and $\epsilon=0.001$, and we have: 
$$ \frac{4-3p}{(2-p)^2} = 1.075$$
$$\frac{4-3p-2r-2\epsilon}{(2-p-r)^2} = 1.10$$
The above shows the addition of a new feature can worsen the fairness of the accuracy optimal classifier.
\end{example}

}

\end{document}